\documentclass{article}

\PassOptionsToPackage{numbers, compress}{natbib}

\usepackage[preprint]{neurips_2026}

\usepackage[utf8]{inputenc} 
\usepackage[T1]{fontenc}    
\usepackage{hyperref}       
\usepackage{url}            
\usepackage{booktabs}       
\usepackage{amsfonts}       
\usepackage{nicefrac}       
\usepackage{microtype}      
\usepackage{xcolor}         

\usepackage{booktabs}   
\usepackage{multirow}   
\usepackage{graphicx}   
\usepackage{hyperref}   
\usepackage{pifont}     
\usepackage{longtable}
\usepackage{float}
\usepackage{booktabs}
\usepackage{tabularx}
\usepackage{amsmath}

\newcommand{\cmark}{\ding{51}}
\newcommand{\xmark}{\ding{55}}

\usepackage{graphicx} 
\usepackage{wrapfig}

\title{Exposing Blind Spots in\\Deep Imbalanced Regression Evaluation}

\author{%
\quad Noah C. Puetz$^{1,2}$ \thanks{Contact: noah\_christoph.puetz@th-koeln.de} \quad Jens U. Brandt$^{1,2}$ \quad \textbf{Marc Hilbert}$^{2,3}$ \\ \quad \textbf{Elena Raponi}$^2$ \quad \textbf{Thomas Bäck}$^2$ \quad \textbf{Thomas Bartz-Beielstein}$^1$\\
$^1$TH Köln \quad $^2$Leiden University \quad $^3$Toyota Racing \\
}

\begin{document}

\maketitle

\begin{abstract}
Deep Imbalanced Regression (DIR) addresses a common failure mode of regression models: target distributions are highly non-uniform, causing models to perform best in densely populated target regions even when reliable performance is required across the full target range. Despite rapid methodological progress, DIR evaluation remains constrained by three blind spots: it is dominated by image-based benchmarks, its standard many-/medium-/few-shot protocol is diagnostic but not decision-complete, and tail-region stability across random seeds has not been systematically evaluated. We revisit DIR evaluation along these three axes. First, we broaden the data domain by evaluating DIR on a multimodal virtual sensing benchmark (\textsc{MuViS}) with nine time-series extrinsic regression tasks across six physical domains, where rare target values often correspond to operationally meaningful regimes. Second, we adopt balanced MAE (\emph{bMAE}) and introduce balanced Mean Absolute Scaled Error (\emph{bMASE}), a scale-normalized metric for decision-complete comparison across methods and datasets. Third, through a repeated reevaluation of six representative DIR methods across multiple random seeds, we show that the tail regions targeted by DIR exhibit particularly high sensitivity to seed-level variability. Our results show that standard virtual-sensing models exhibit substantial tail degradation hidden by global MAE, that existing DIR methods can improve balanced performance but transfer unevenly to multimodal time-series data, and that tail-region instability remains a largely hidden failure mode under current DIR evaluation practice. Together, these findings and our publicly available code provide a reproducible basis for future DIR research toward regression systems that capture rare target regimes as reliably as common ones.
\end{abstract}

\begin{figure}[ht]
\centering
\vspace{-15pt}
\includegraphics[width=\textwidth]{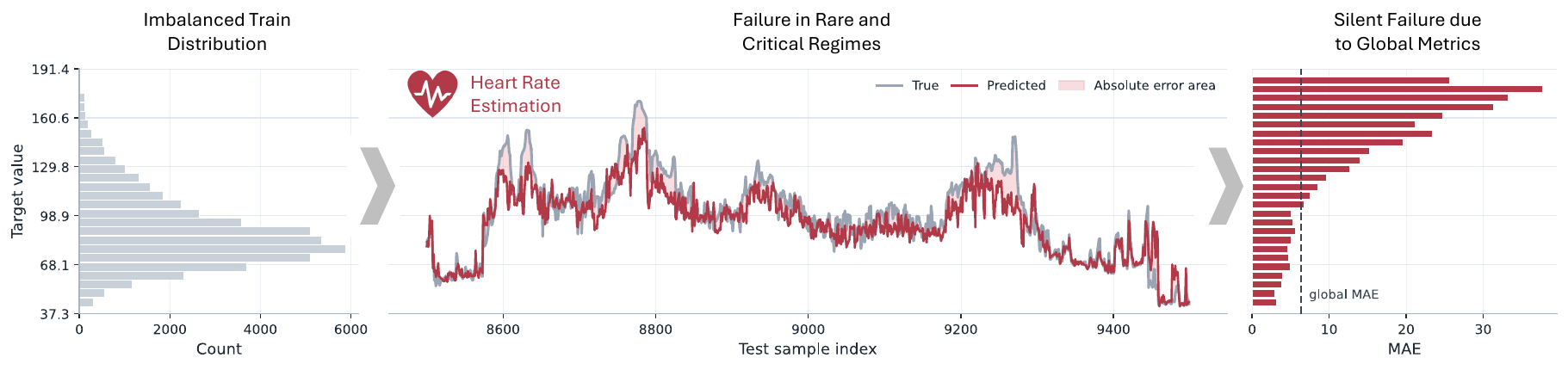}
\vspace{-15pt}
\caption{\textbf{Motivating Example}. In heart-rate estimation \citep{reiss_deep_2019}, the training distribution is strongly imbalanced (left), prediction quality degrades in rare but critical regimes (middle), and this failure is only weakly reflected by the global MAE (right).}
\label{fig:graphical_abstract}
\vspace{-5pt}
\end{figure}

\section{Introduction}
\label{sec:1}
In many real-world regression problems, the target variable is highly imbalanced. Some regions of the target space are densely populated with training samples, whereas others, often the most extreme and practically most consequential values, are observed only rarely \citep{yang_delving_2021, puetz_deconstructing_2026}. When trained on such data, standard deep learning models naturally achieve their best predictive accuracy in well-supported regions of the training distribution and degrade in underrepresented ones \citep{yang_delving_2021, shalev-shwartz_understanding_2014, zhang_physics-based_2026}. In many applications, however, predictive performance is required not only where data are abundant, but across the entire target range \citep{ren_balanced_2022}. 
\autoref{fig:graphical_abstract} provides such an example: In heart rate estimation most of the data is collected during normal operation of the human body \cite{reiss_deep_2019}, generating an imbalanced train distribution and leading to poor performance in the heart rate spikes, which ultimately results in worst performance in the most critical regimes. In such a scenario, the imbalanced training distribution becomes misaligned with the desired evaluation objective: rare outcomes receive little influence during optimization, despite being equally important at test time. Deep Imbalanced Regression (DIR) \citep{yang_delving_2021} addresses this mismatch and characterizes it as an implicit target distribution shift \citep{puetz_deconstructing_2026}. 

Motivated by this problem, recent years have seen a growing body of DIR methods that aim to improve performance in under-represented regions of the target space \citep{puetz_deconstructing_2026}. However, despite this methodological progress, the existing DIR evaluation design still suffers from three important blind spots.  First, the \textit{Data Blind Spot}: While DIR claims to be data domain agnostic its empirical ecosystem is still overwhelmingly concentrated in computer vision, while other domains like time-series settings remain largely absent (see \autoref{app:data_blindspot})
\citep{puetz_deconstructing_2026}.
Second, the \textit{Metric Blind Spot}: the established reporting practice in DIR is to partition the target space into many-, medium-, and few-shot regions and report separate error values for each \citep{yang_delving_2021, puetz_deconstructing_2026}. However, these partitions are arbitrary in continuous target spaces, and are difficult to interpret consistently across datasets \citep{liu_large-scale_2019}. More importantly, they do not produce a single aggregate quantity for ranking methods. As a result, comparisons remain partly subjective: a method may improve in underrepresented areas at the expense of well-represented ones, yet the field lacks a principled way to assess whether this trade-off is beneficial. 
Third, the \textit{Stability Blind Spot}: current DIR evaluation pays limited attention to experimental stability. Methods are frequently compared using single runs or only a small number of repetitions, even though sparse supervision in the distribution tails potentially makes these regions especially vulnerable to variance from random initialization and optimization noise \citep{bouthillier_accounting_2021}. If DIR's central claim is improved performance in underrepresented regions, then stability in those regions is part of what needs to be evaluated. 

In this work, we argue that meaningful DIR evaluation must go beyond the evaluation and reporting conventions that currently dominate the field. Therefore aligned with the three blind spots and supported by our motivating example in \autoref{fig:graphical_abstract} we revisit DIR from three complementary angles:

\begin{enumerate}
    \item Addressing the \textit{Data Blind Spot}, we broaden the DIR data domain to \textsc{MuViS} (Multimodal Virtual Sensing Benchmark) \cite{brandt_muvis_2026}, a collection of multimodal time-series extrinsic regression tasks specialized for virtual sensing and therefore characterized by a clear domain-level requirement for reliable performance across the full target range. 
    
    \item Addressing the \textit{Metric Blind Spot}, we formalize a macro-averaged, distribution-invariant view of DIR evaluation through \textit{bMAE} \cite{ren_balanced_2022} and introduce the novel, scale-invariant \textit{bMASE}. We contrast the prevailing many-/medium-/few-shot reporting with standard global regression metrics such as mean absolute error (MAE), as well as with these balanced alternatives, and show that the bMASE creates an objective cross-method and cross-dataset ranking, that is not doable with the current evaluation design.

    \item Addressing the \textit{Stability Blind Spot}, we run repeated experiments to probe seed-level performance variability in DIR methods, focusing on whether underrepresented target regions exhibit disproportionately high performance variance.
\end{enumerate}

\section{Related Work}
\label{sec:2}
The problem of imbalanced training data is not new in deep learning, but most imbalance-aware methods were developed for classification, where the output space is categorical and finite~\citep{he_learning_2009}. Extending these ideas to continuous and potential infinite target spaces is non-trivial. \citet{yang_delving_2021} introduced DIR as a distinct problem setting and showed that reweighting and resampling strategies transferred from classification perform poorly once the notion of a class boundary disappears. Since then, DIR has developed into a substantial methodological literature, with existing methods broadly grouped into three families \citep{puetz_deconstructing_2026}: \emph{algorithm-level} methods that modify the loss or sample weighting (LDS~\citep{yang_delving_2021}, SqInv~\citep{yang_delving_2021}, Focal-$L_1$~\citep{yang_delving_2021}, Balanced MSE~\citep{ren_balanced_2022}, DenseLoss~\citep{steininger_density-based_2021}, VIR~\citep{wang_variational_2023}, Dist~Loss~\citep{nie_dist_2025}); \emph{representation-learning} methods that reshape the latent space (FDS~\citep{yang_delving_2021}, RankSim~\citep{gong_ranksim_2022}, Ordinal Entropy~\citep{zhang_improving_2023}, ConR~\citep{keramati_conr_2023}, RnC~\citep{zha_rank-n-contrast_2023}, Geometric Representation Constraints~\citep{dong_improve_2025}, PRIME~\citep{lim_prime_2025}); and \emph{architectural} methods that modify the predictor itself (UVote~\citep{jiang_uncertainty_2024}, HCA~\citep{xiong_deep_2024}, Multi-Classification~\citep{lin_let_2024}, IM-Context~\citep{nejjar_im-context_2024}). Additionally, these methods can be divided into \emph{orthogonal} approaches that compose with standard regression pipelines and \emph{non-orthogonal} approaches that require specific architectures or substantially restructured training~\citep{puetz_deconstructing_2026}.

Despite DIR's methodological progress, its empirical basis remains narrow. Across the fifteen DIR method papers surveyed by \citet{puetz_deconstructing_2026}, fourteen paper evaluate on at least one image-based benchmark, predominantly \textsc{AgeDB-DIR} \citep{yang_delving_2021, moschoglou_agedb_2017} or \textsc{IMDB-WIKI-DIR} \citep{yang_delving_2021, rothe_deep_2018}; eight additionally evaluate on text regression (\textsc{STS-B-DIR} \citep{yang_delving_2021, cer_semeval-2017_2017}); four on synthetic or tabular data; and only three on time-series data (full table shown in \autoref{app:data_blindspot}). Moreover, each of these three evaluations use a different dataset with no overlap between papers (\textsc{SHHS-DIR} \citep{quan_sleep_1997} in \citet{yang_delving_2021}, \textsc{TUAB} \citep{engemann_reusable_2022} in \citet{zha_rank-n-contrast_2023}, and \textsc{ECG-K-DIR} \citep{johnson_mimic-iv_2023} in \citet{nie_dist_2025}) which prevents direct cross-method comparison. None of the three releases a public training pipeline for its respective time-series task, and two are difficult to reconcile with the continuous-target, deep-regression premise that motivates DIR: \textsc{SHHS-DIR} provides with $1{,}892$ a limited amount of training samples with only $21$ discrete target values, and \textsc{TUAB} provides $1{,}246$ samples with integer targets in $[0,95]$. As a result, current DIR evidence says far more about image benchmarks than about the broader class of imbalanced continuous prediction problems the field claims to address. A more complete evaluation landscape should include data domains that differ structurally from images while preserving the core DIR challenge: continuous targets with sparsely supported but practically relevant regions. 

\section{Extending the Evaluation Framework of DIR}
\label{sec:framework}

We extend DIR evaluation along the three blind spots identified in \autoref{sec:1}. First, in \autoref{sec:3_1} we broaden the data domain by evaluating on multimodal time-series regression. Second, in \autoref{sec:3_2} we replace many-/medium-/few-shot reporting with balanced scalar metrics that are aligned with continuous targets and enable direct comparison across datasets. Third, we treat stability in underrepresented regions as an evaluation criterion and analyze it empirically in \autoref{sec:4_3}.

\subsection{Benchmark Extension: \textsc{MuViS-DIR}}
\label{sec:3_1}
\begin{figure}[t]
\centering
\includegraphics[width=0.99\textwidth]{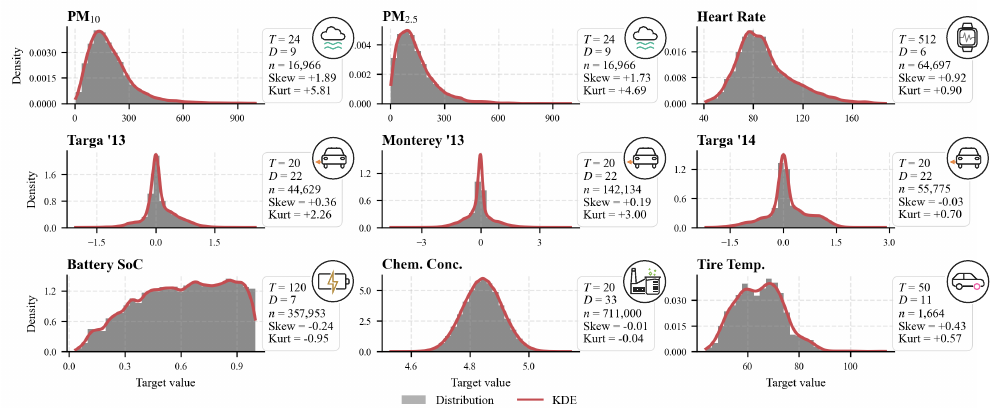}
\vspace{-5pt}
\caption{
Empirical target distributions of the nine \textsc{MuViS-DIR} tasks, illustrating task-specific target imbalance across continuous sensor-like quantities. Legends report sequence length $T$, feature channels $D$, sample size $n$, skewness, and kurtosis. Domain icons follow \citet{brandt_muvis_2026}.
}
\vspace{-10pt}
\label{fig:target_distributions}
\end{figure}

To contribute to closing the \textit{Data Blind Spot}, we use \textsc{MuViS}~\citep{brandt_muvis_2026}, a multimodal virtual sensing benchmark with nine continuous regression tasks across six physical domains. Virtual sensing denotes the data-driven inference of hard-to-measure physical quantities from available sensor measurements \cite{muir_virtual_1990, martin_virtual_2021}. The task is sequence-to-value regression, at the intersection of virtual sensing and multimodal dynamic time-series learning~\citep{mohapatra_maestro_2025}, with an interface to time-series extrinsic regression (TSER)~\citep{tan_monash_2020}. Combining DIR and \textsc{MuViS} yields the following problem setting for \textsc{MuViS-DIR}:

\paragraph{Problem setting of \textsc{MuViS-DIR}:}
\label{problem_setting}
We consider a dataset $\mathcal{D}=\{(\mathcal{X}_i,y_i)\}_{i=1}^{N}$, where each input $\mathcal{X}_i=(x_i^1,\ldots,x_i^M)$ consists of $M$ modality-specific time series with $x_i^j\in\mathbb{R}^{D^j\times T^j}$. Modality $j$ provides $D^j$ feature channels over $T^j$ time steps, and $y_i\in\mathbb{R}$ denotes the continuous sensor-like target value $y_i(t_0)$ at reference time $t_0$. A deep model $f_\theta$ maps the multimodal sequential input to an estimate $\hat{y}_i=f_\theta(\mathcal{X}_i)$. In the DIR setting, the empirical target distribution $p_{\mathrm{train}}(y)$ is highly non-uniform: some target intervals are well-represented, whereas others contain few or no samples. The evaluation objective, however, is to generalize to a more balanced desired evaluation distribution $p_{\mathrm{eval}}(y)$. This mismatch induces an implicit target-distribution shift between training and evaluation, weakening the standard empirical-risk-minimization (ERM) assumption that training and evaluation samples are drawn from the same target distribution. 

\textsc{MuViS-DIR} is a useful DIR testbed because it combines continuous targets, heterogeneous sequential inputs, and domain-level reasons to care about rare target values. In many virtual sensing tasks, underrepresented regions correspond to safety-critical \citep{zhang_cautionary_2017, chen_evidence_2013}, clinically relevant \citep{reiss_deep_2019}, or physically extreme operating \citep{downs_plant-wide_1993, zhang_physics-based_2026} regimes rather than merely rare labels. Poor tail performance is therefore a practical failure mode of the trained model. Figure~\ref{fig:target_distributions} visualizes the diverse target distributions used in our study; full dataset statistics, domain motivations for reliable performance across the target range, and imbalance summaries are reported in \autoref{app:data_blindspot}.

\subsection{Metric Extension: Balanced Scalar Evaluation}
\label{sec:3_2}
\begin{wrapfigure}{r}{0.45\textwidth}
    \centering
    \includegraphics[width=\linewidth]{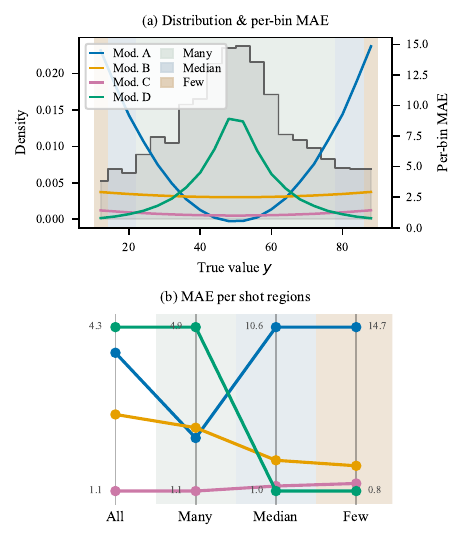}
    \vspace{-10pt}
    \caption{Failure mode of many-/medium-/few-shot reporting. 
(a) Four synthetic predictors exhibit different per-bin error profiles over the same imbalanced target distribution. 
(b) Region-wise MAE produces different rankings depending on the selected shot region and therefore does not provide a unique model comparison criterion.}
    \label{fig:yang_critic}
    \vspace{-20pt}
\end{wrapfigure}

The dominant DIR protocol of \citet{yang_delving_2021} reports MAE separately over many-, medium-, and few-shot regions, which creates two problems for continuous regression. First, the thresholds are inherited from long-tailed classification \citep{liu_large-scale_2019} (few-shot: $<20$ samples, many-shot: $>100$ samples, medium-shot: otherwise) and are arbitrary once the target space is continuous. Second, the protocol produces three region-specific scores but no scalar ranking criterion. A model may improve rare regions while degrading dense ones, yet the protocol provides no principled rule for deciding whether this trade-off is favorable. The following motivating example illustrates this ambiguity.

\paragraph{Motivating example.} \autoref{fig:yang_critic}~(a) shows four synthetic predictors with different error profiles over the same imbalanced target distribution. Model~A resembles the typical ERM failure mode \citep{puetz_deconstructing_2026}: low error in the dense head and increasing error in the tails. Model~C is nearly uniform across the target range and is visually the most robust predictor under a balanced evaluation objective. Model~D exposes the ambiguity most clearly: it performs worst in the dense head but best in the medium- and few-shot regions, precisely the regimes DIR is meant to improve. Depending on the evaluation preference, Model~D can therefore be viewed either as a poor model that sacrifices common cases or as a strong DIR model that improves rare regimes (\autoref{fig:yang_critic}~(b)). The shot-based protocol reports this trade-off but does not resolve it. A naive average over the three shot scores would introduce another arbitrary decision rule, ranking Model~D ahead of Model~B (mean MAE: 2.13 vs. 2.88) without justifying how tail gains should be weighted against head degradation. Thus, the protocol can be diagnostic but not decision-complete: it reveals (arbitrary) regional behavior, yet it does not define an objective comparison criterion.

Motivated by this limitation, and by the premise that DIR evaluation should reward models that perform consistently well across an idealized balanced $p_{\mathrm{eval}}(y)$ \citep{puetz_deconstructing_2026}, we adopt balanced scalar metrics that weight target regions uniformly rather than by sample density. Since $y$ is continuous, we follow~\citet{ren_balanced_2022} and approximate uniform weighting over the target range by discretizing $[a,b]$ into $K$ equal-width bins; \autoref{app:cont-vs-disc} verifies the agreement between the continuous and discretized formulations for sufficiently large test sets. 

\paragraph{Balanced MAE.}\label{sec:bmae} We macro-average the per-bin MAE over the $K$ equal-width bins \citep{ren_balanced_2022}:
\begin{equation}\label{eq:bmae}
  \mathrm{bMAE}
  \;=\;
  \frac{1}{K}\sum_{k=1}^{K}\,
  \frac{1}{n_{k}}\!\sum_{i\in\mathrm{bin}_{k}}\!|y_{i}-\hat{y}_{i}|\,,
\end{equation}
where $n_{k}$ is the number of test samples in bin~$k$. Every bin receives equal weight regardless of sample count, reducing the dependence on the test-set density while preserving the interpretability of MAE in the original target units.

\paragraph{Balanced MASE.} While bMAE provides an interpretable absolute error, it is scale-dependent and cannot be compared across datasets. Because the nine datasets in our benchmark span fundamentally different physical quantities and scales, we introduce the balanced Mean Absolute Scaled Error (bMASE, based on the mean absolute scaled error by \citet{hyndman_another_2006}), which normalises bMAE by the error of a trivial constant predictor:
\begin{equation}\label{eq:bmase}
  \mathrm{bMASE}
  \;=\;
  \frac{\frac{1}{K}\sum_{k=1}^{K}\,\frac{1}{n_{k}}\!\sum_{i\in\mathrm{bin}_{k}}\!\vert y_{i}-\hat{y}_{i}\vert}
  {\frac{1}{K}\sum_{k=1}^{K}\,\frac{1}{n_{k}}\!\sum_{i\in\mathrm{bin}_{k}}\! \vert y_{i}-c_{\mathrm{ref}}\vert}
  \;=\;
  \frac{\mathrm{bMAE}(f_{\theta})}{\mathrm{bMAE}(c_{\mathrm{ref}})}
\end{equation}
where $c_{\mathrm{ref}} = \mathrm{median}(y_{\mathrm{train}})$ is
the $L_{1}$-optimal constant predictor.
A $\mathrm{bMASE} < 1$ indicates improvement over the trivial
predictor; values closer to~$0$ indicate stronger performance.
Because $c_{\mathrm{ref}}$ normalises out the dataset-specific error
scale, bMASE values are directly comparable across datasets and can be
aggregated via the geometric mean:
\begin{equation}\label{eq:bmase-agg}
  \overline{\mathrm{bMASE}}
  \;=\;
  \Biggl(\,\prod_{d=1}^{\mathrm{D}}\,
    \frac{\mathrm{bMAE}_{d}(f_{\theta})}
         {\mathrm{bMAE}_{d}(c_{\mathrm{ref}})}
  \Biggr)^{\frac{1}{\mathrm{D}}}
\end{equation}
The geometric mean is the only aggregation of normalised ratios that
is invariant to the choice of reference, i.e.\ the ranking is
identical whether one normalises by the trivial predictor, the best
model, or any other baseline~\citep{fleming_how_1986}.

We use $L_1$-based metrics because balanced evaluation gives equal weight to every target bin, including bins with very small sample counts. In such bins, squared-error metrics are especially unstable: per-bin MSE depends on fourth-order error moments, whereas MAE depends only on second-order moments through the variance of $|e_i|$, with $e_i=y_i-\hat{y}_i$~\citep{huber_robust_1964}. Since metric stability in rare target regions is central to our setting, we use the $L_1$-based bMASE rather than balanced variants of RMSE or $R^2$.

Both \autoref{eq:bmae} and \autoref{eq:bmase} require a target range $[a,b]$ and a number of bins $K$. We set $a=\min(y_{\mathrm{test}})$ and $b=\max(y_{\mathrm{test}})$, and choose $K$ using the Freedman--Diaconis rule~\citep{freedman_histogram_1981},
\[
w_{\mathrm{FD}} = 2\,\mathrm{IQR}(y_{\mathrm{test}})\,N_{\mathrm{test}}^{-1/3},
\qquad
K = \left\lceil\frac{b-a}{w_{\mathrm{FD}}}\right\rceil .
\]
Thus, the binning is deterministic, dataset-specific, shared by bMAE and bMASE, and not treated as a tunable hyperparameter.

\section{Experiments}
\label{sec:4}
We organize the experiments around the three evaluation blind spots introduced in \autoref{sec:1}. First, we test whether the DIR failure mode appears in \textsc{MuViS} (\autoref{sec:4_1}). Second, we evaluate whether established DIR methods improve balanced performance under bMAE and bMASE (\autoref{sec:4_2}). Third, we analyze whether model performance in the same low-density regions is also less stable across random seeds (\autoref{sec:4_3}). Code for reproducing the experiments is available at \url{www.github.com/noah-puetz/muvis-dir}.

\subsection{Imbalance Bias in Standard Virtual Sensors}
\label{sec:4_1}

We first examine whether \textsc{MuViS-DIR} exhibits the imbalance bias that motivates DIR \citep{yang_delving_2021}. For each dataset, we train the dataset-specific \textsc{ResNet1D} architectures provided by \citet{brandt_muvis_2026} with an $L_1$ loss across ten random seeds. This provides a baseline for the later DIR comparison while avoiding architectural confounds. Results for the remaining \textsc{MuViS} baselines under bMAE and bMASE are reported in \autoref{app:reevaluation}; among them, \textsc{ResNet1D} obtains the best aggregate bMASE and is therefore used as the basis for the main analysis.

\begin{figure}[t]  
\centering  
\includegraphics[width=\linewidth]{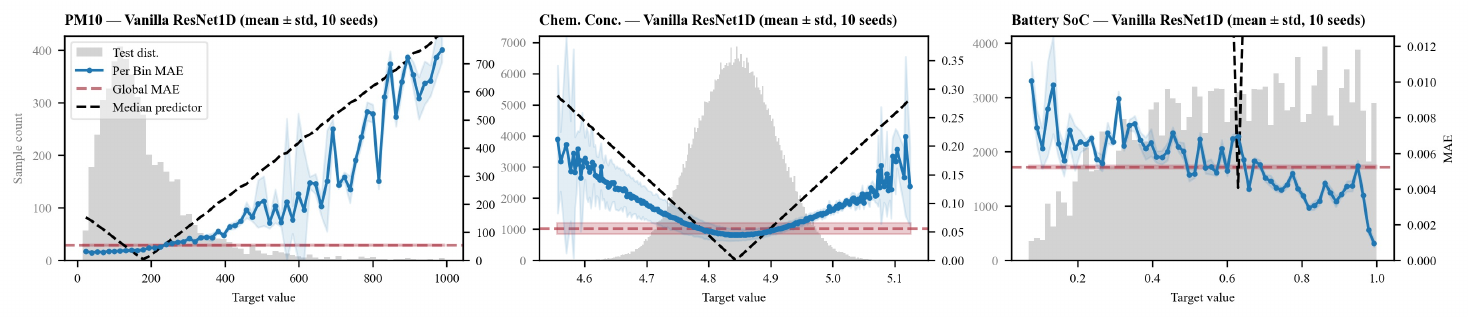}  
\vspace{-20pt}
\caption{  
Per-bin MAE of \textsc{ResNet1D} across three representative tasks from \textsc{MuViS} \citep{brandt_muvis_2026}. Error increases sharply in sparse target regions; on \textsc{PM10}, the extreme upper tail exceeds the median-predictor baseline.  
}  

\label{fig:resnet-tail-bias}  
\end{figure}

\autoref{fig:resnet-tail-bias} shows that the core DIR failure mode is present in \textsc{MuViS}. Across the representative tasks, per-bin MAE increases in sparsely supported target regions, with the single exception of the per bin MAE on \textsc{Battery SoC}. \textsc{Battery SoC} has the least imbalanced target support among the shown tasks, and its per-bin MAE remains comparatively stable across the target range. Table~\ref{tab:bmae-vs-mae} shows that the failure mode on \textsc{PM10} is largely hidden by global MAE. Only when target regions are weighted equally by the \textit{bMAE}, the \textsc{ResNet1D} obtains a $4.42\times$ increase. This illustrates why balanced metrics across target range are practically important: predictions on \textsc{PM10} in the extreme upper tail, $y>600$ approach the performance of a naive median predictor, but these rare target ranges are also corresponding to critical air-quality regimes which would trigger public health advisories \citep{chen_evidence_2013}. The MAE let the model appear competitive, yet it fails in the operating region where reliability is most consequential (similar masking effect appears on \textsc{PM2.5} and \textsc{Monterey}, see \autoref{app:performance_comparison}).

The many-/median-/few-shot protocol is less reliable in this setting. Some datasets contain no bins satisfying a given shot criterion, producing undefined entries, and the fixed thresholds can still overweight denser subregions inside the nominal few-shot range. On \textsc{PM2.5}, the few-shot region spans approximately $y=300$ to $y=1000$, but most samples within that region are concentrated near its lower end. Consequently, few-shot MAE underweights the most severe upper-tail errors, while bMAE exposes them. Thus, the shot-based reporting, meant for reporting models sensitivity due to data imbalance, can itself remain sensitive to within-region imbalance.

\begin{table}[t]  
\centering  
\caption{  
Contrasting classical MAE, many-/median-/few-shot MAE, and bMAE for each dataset with its corresponding \textsc{ResNet1D} configuration. Mean $\pm$ 95\% bootstrap CI half-width on the mean over 10 seeds. 
The MAE vs. bMAE ratio quantifies how strongly global MAE underestimates balanced target-range error. Larger values indicate stronger imbalance-induced masking.  
}  
\label{tab:bmae-vs-mae}  
\resizebox{\textwidth}{!}{
\begin{tabular}{lrrrrrr}
\toprule
Metric & MAE & Many-MAE & Medi.-MAE & Few-MAE & bMAE & MAE vs.\ bMAE \\
Dataset &  &  &  &  &  &  \\
\midrule
Battery SoC & 0.0052\scriptsize{$\pm$ 0.0001} & 0.0052\scriptsize{$\pm$ 0.0001} & -- & -- & 0.0056\scriptsize{$\pm$ 0.0001} & 1.0636 \\
Chem. Conc. & 0.0556\scriptsize{$\pm$ 0.0058} & 0.0554\scriptsize{$\pm$ 0.0059} & 0.1245\scriptsize{$\pm$ 0.0111} & 0.1480\scriptsize{$\pm$ 0.0164} & 0.0931\scriptsize{$\pm$ 0.0096} & 1.6747 \\
Heart Rate & 4.1265\scriptsize{$\pm$ 0.1135} & 3.9351\scriptsize{$\pm$ 0.1084} & 6.3535\scriptsize{$\pm$ 0.1997} & 5.7599\scriptsize{$\pm$ 0.4316} & 5.1623\scriptsize{$\pm$ 0.1549} & 1.2510 \\
Monterey & 0.1289\scriptsize{$\pm$ 0.0320} & 0.1113\scriptsize{$\pm$ 0.0244} & 0.2060\scriptsize{$\pm$ 0.0611} & 0.4445\scriptsize{$\pm$ 0.1958} & 0.4377\scriptsize{$\pm$ 0.1820} & 3.3955 \\
PM10 & 54.6721\scriptsize{$\pm$ 1.7079} & 38.0200\scriptsize{$\pm$ 0.5720} & 83.9907\scriptsize{$\pm$ 2.3227} & 291.9299\scriptsize{$\pm$ 36.5234} & 242.0897\scriptsize{$\pm$ 16.9107} & 4.4280 \\
PM2.5 & 38.3933\scriptsize{$\pm$ 2.1100} & 28.5859\scriptsize{$\pm$ 1.3722} & 84.5718\scriptsize{$\pm$ 5.8982} & 159.7854\scriptsize{$\pm$ 13.3461} & 185.3354\scriptsize{$\pm$ 9.5653} & 4.8273 \\
Targa '13 & 0.0505\scriptsize{$\pm$ 0.0005} & 0.0468\scriptsize{$\pm$ 0.0004} & 0.0639\scriptsize{$\pm$ 0.0015} & 0.1109\scriptsize{$\pm$ 0.0048} & 0.0950\scriptsize{$\pm$ 0.0039} & 1.8812 \\
Targa '14 & 0.0725\scriptsize{$\pm$ 0.0134} & 0.0717\scriptsize{$\pm$ 0.0119} & 0.0788\scriptsize{$\pm$ 0.0241} & 0.0837\scriptsize{$\pm$ 0.0256} & 0.0777\scriptsize{$\pm$ 0.0214} & 1.0720 \\
Tire Temp. & 3.1058\scriptsize{$\pm$ 0.2498} & -- & 2.9534\scriptsize{$\pm$ 0.2342} & 4.8940\scriptsize{$\pm$ 0.4510} & 3.4308\scriptsize{$\pm$ 0.2748} & 1.1046 \\
\bottomrule
\end{tabular}
}
\end{table}

\subsection{DIR Methods under Balanced Evaluation}

\label{sec:4_2}
\begin{table}[t]  
\centering  
\caption{  
Mean bMASE $\pm$ 95\% bootstrap CI half-width over 10 seeds for each method--dataset pair. Below \textsc{Vanilla} are italicized; Best value in each row is bold. GMean aggregates across datasets via geometric mean. Lower is better; $\mathrm{bMASE}<1$ improves over the trivial median predictor.  
}  
\label{tab:dir-results}  
\resizebox{\textwidth}{!}{
\begin{tabular}{lrrrrrrr}
\toprule
Method & Vanilla & ConR & Focal-$L_1$ & LDS & RnC & SQInv & UVote \\
Dataset &  &  &  &  &  &  &  \\
\midrule
Battery SoC & 0.0229\scriptsize{$\pm$ 0.0003} & 0.0306\scriptsize{$\pm$ 0.0089} & 0.0280\scriptsize{$\pm$ 0.0051} & \textbf{\textit{0.0224}}\scriptsize{$\pm$ 0.0004} & 0.2536\scriptsize{$\pm$ 0.0977} & 0.0262\scriptsize{$\pm$ 0.0053} & 0.0251\scriptsize{$\pm$ 0.0008} \\
Chem. Conc. & 0.6910\scriptsize{$\pm$ 0.0709} & \textit{0.6485}\scriptsize{$\pm$ 0.0483} & \textit{0.6343}\scriptsize{$\pm$ 0.0134} & \textbf{\textit{0.5826}}\scriptsize{$\pm$ 0.0325} & \textit{0.6349}\scriptsize{$\pm$ 0.0197} & \textit{0.6371}\scriptsize{$\pm$ 0.0673} & \textit{0.6056}\scriptsize{$\pm$ 0.0028} \\
Heart Rate & 0.1277\scriptsize{$\pm$ 0.0038} & 0.1925\scriptsize{$\pm$ 0.0921} & \textbf{\textit{0.0748}}\scriptsize{$\pm$ 0.0028} & \textit{0.1130}\scriptsize{$\pm$ 0.0074} & \textit{0.0923}\scriptsize{$\pm$ 0.0052} & \textit{0.1232}\scriptsize{$\pm$ 0.0234} & \textit{0.0782}\scriptsize{$\pm$ 0.0039} \\
Monterey & 0.2511\scriptsize{$\pm$ 0.1028} & 0.2695\scriptsize{$\pm$ 0.1162} & \textit{0.2215}\scriptsize{$\pm$ 0.0750} & \textit{0.2401}\scriptsize{$\pm$ 0.0962} & 0.3622\scriptsize{$\pm$ 0.2632} & \textit{0.2260}\scriptsize{$\pm$ 0.0907} & \textbf{\textit{0.0853}}\scriptsize{$\pm$ 0.0016} \\
PM10 & 0.6869\scriptsize{$\pm$ 0.0483} & \textit{0.6793}\scriptsize{$\pm$ 0.0258} & \textit{0.6171}\scriptsize{$\pm$ 0.0094} & \textbf{\textit{0.4765}}\scriptsize{$\pm$ 0.0254} & \textit{0.6663}\scriptsize{$\pm$ 0.0085} & \textit{0.6230}\scriptsize{$\pm$ 0.0214} & \textit{0.6792}\scriptsize{$\pm$ 0.0353} \\
PM2.5 & 0.5705\scriptsize{$\pm$ 0.0286} & \textit{0.5507}\scriptsize{$\pm$ 0.0289} & \textit{0.5242}\scriptsize{$\pm$ 0.0123} & \textit{0.5489}\scriptsize{$\pm$ 0.0541} & \textit{0.5675}\scriptsize{$\pm$ 0.0227} & \textit{0.5457}\scriptsize{$\pm$ 0.0287} & \textbf{\textit{0.5142}}\scriptsize{$\pm$ 0.0080} \\
Targa '13 & \textbf{0.1116}\scriptsize{$\pm$ 0.0047} & 0.1136\scriptsize{$\pm$ 0.0025} & 0.1933\scriptsize{$\pm$ 0.0091} & 0.1212\scriptsize{$\pm$ 0.0039} & 0.3452\scriptsize{$\pm$ 0.1847} & 0.1132\scriptsize{$\pm$ 0.0029} & 0.1136\scriptsize{$\pm$ 0.0051} \\
Targa '14 & 0.1276\scriptsize{$\pm$ 0.0349} & 0.1319\scriptsize{$\pm$ 0.0409} & 0.1326\scriptsize{$\pm$ 0.0099} & \textit{0.1273}\scriptsize{$\pm$ 0.0283} & 0.1746\scriptsize{$\pm$ 0.0332} & \textit{0.1183}\scriptsize{$\pm$ 0.0195} & \textbf{\textit{0.1055}}\scriptsize{$\pm$ 0.0133} \\
Tire Temp. & 0.3662\scriptsize{$\pm$ 0.0296} & \textit{0.3618}\scriptsize{$\pm$ 0.0230} & 0.3777\scriptsize{$\pm$ 0.0284} & \textit{0.3534}\scriptsize{$\pm$ 0.0261} & 0.3933\scriptsize{$\pm$ 0.0358} & \textit{0.3505}\scriptsize{$\pm$ 0.0221} & \textbf{\textit{0.3313}}\scriptsize{$\pm$ 0.0254} \\
\midrule
GMean & 0.2163\scriptsize{$\pm$ 0.0158} & 0.2338\scriptsize{$\pm$ 0.0272} & \textit{0.2135}\scriptsize{$\pm$ 0.0117} & \textit{0.1997}\scriptsize{$\pm$ 0.0148} & 0.3314\scriptsize{$\pm$ 0.0364} & \textit{0.2084}\scriptsize{$\pm$ 0.0202} & \textbf{\textit{0.1732}}\scriptsize{$\pm$ 0.0057} \\
\bottomrule
\end{tabular}
}
\end{table}

We next ask whether established DIR methods can reduce this imbalance bias. We evaluate six representative and state-of-the-art methods (based on the extrapolation benchmark from \citet{puetz_deconstructing_2026}) covering the main DIR families from \autoref{sec:2}: algorithm-level methods, LDS, SQInv, and Focal-$L_1$; representation-learning methods, ConR and RnC; and the architectural method UVote. LDS, SQInv, Focal-$L_1$, and ConR are orthogonal interventions. All methods are trained on top of the same \textsc{ResNet1D} from \autoref{sec:4_1}, and each method--dataset pair is selected by a hyperparameter sweep (explained in detail in \autoref{app:repro:selection}).

\begin{wrapfigure}{r}{0.36\textwidth}
\centering  
\vspace{-15pt}
\includegraphics[width=\linewidth]{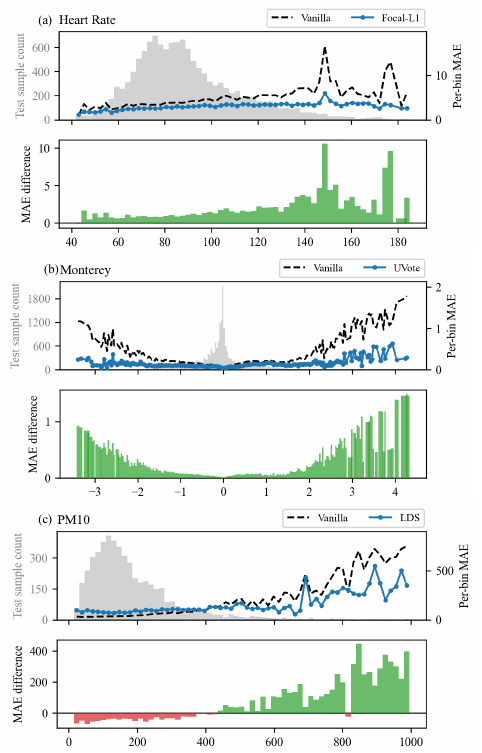}  
\vspace{-20pt}
\caption{  
Per-bin improvements over \textsc{Vanilla}. Top row compares per-bin MAE; bottom row shows the signed difference to the baseline. (Full Overview: \autoref{fig:appendix_method_vs_vanilla})
}  
\vspace{-20pt}
\label{fig:dir-improvements}  
\end{wrapfigure}

\autoref{tab:dir-results} demonstrates the advantage of bMASE as a single scalar comparison criterion: it provides a cross-method and cross-dataset ranking that many-/medium-/few-shot reporting cannot supply.  Four out of six methods improve the vanilla baseline, showing that imbalance correction transfers to multimodal virtual sensing. However, the gains are less regular than in common computer-vision DIR benchmarks. UVote achieves the best aggregate score, reducing GMean bMASE from $0.2163$ to $0.1732$, and is best on four datasets. LDS is the strongest evaluated orthogonal method, reaching $0.1997$ GMean and producing the largest single improvement on \textsc{PM10}, from $0.6869$ to $0.4765$. SQInv and Focal-$L_1$ also improve the aggregate score, whereas RnC substantially degrades performance.

\autoref{fig:dir-improvements} illustrates how these aggregate gains arise. Focal-$L_1$ improves high-heart-rate regions on \textsc{PPGDalia} and UVote substantially reduces tail error on \textsc{Monterey} with both achieving smaller improvements across the whole respective target range. In contrast LDS compromises in the well-represented region for better tail-performance on \textsc{PM10}.

These results support two conclusions. First, the relationship between \textsc{MuViS} and DIR is bidirectional: \textsc{MuViS} exposes how well DIR methods transfer to multimodal time-series regression, while DIR methods yield meaningful gains in underrepresented sensor regimes. Second, direct transfer from existing DIR methods remains incomplete. Strong methods in image-based DIR do not necessarily dominate multimodal time-series regression, making \textsc{MuViS-DIR} a realistic benchmark for future method development.

\subsection{Tail Instability Across Seeds}
\label{sec:4_3}
\begin{figure}[t]  
\centering  
\includegraphics[width=\linewidth]{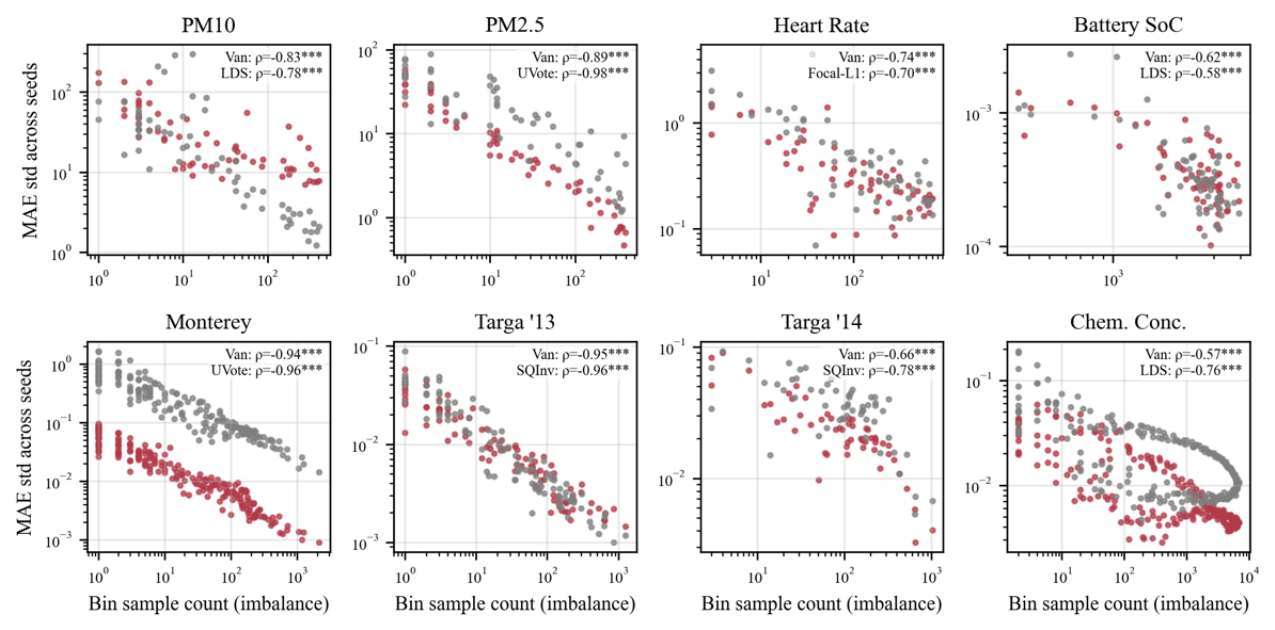}\vspace{-10pt}
\caption{  
Relationship between local target support and seed-level instability. Each point is one bin; the $x$-axis shows bin sample count ($\leftarrow$ less data) and the $y$-axis shows the standard deviation of per-bin MAE across ten seeds ($\uparrow$ higher instability). Both axes are log-scaled. Gray denotes \textsc{Vanilla}; red denotes the best-performing DIR method according to \autoref{tab:dir-results} per dataset.  
}  
\vspace{-10pt}
\label{fig:stability-density}  
\end{figure}

Finally, we evaluate the \textit{Stability Blind Spot}. Since DIR methods are designed to improve low-density regions, their performance should be reproducible precisely in those regions. Motivated by prior work on random-seed variability in deep learning~\citep{reimers_reporting_2017, colas_how_2018}, we compute the standard deviation of per-bin MAE across ten seeds for every method--dataset pair. Rather than treating ten seeds as a sufficiency guarantee, we use this repeated-run setting to test whether instability is already visible in the target regimes DIR aims to improve.

Figure~\ref{fig:stability-density} shows, that across datasets, vanilla \textsc{ResNet1D} becomes less stable as local target support decreases, with Spearman correlations between bin count and per-bin standard deviation ranging from $-0.57$ to $-0.95$. The same trend persists for the best-performing DIR methods: each method either shows an even stronger density--instability relation or differs from vanilla by less than $0.05$ correlation points.

\begin{wrapfigure}{r}{0.48\textwidth}
\vspace{-25pt}
\centering  
\includegraphics[width=\linewidth]{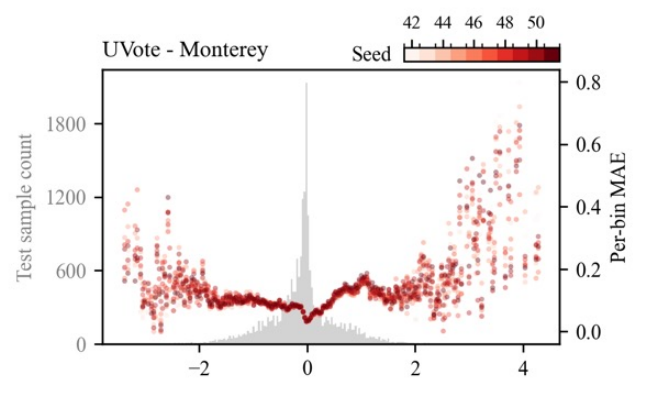}  
\vspace{-20pt}
\caption{  
Seed-level instability for UVote on \textsc{Monterey}. Each point corresponds to one seed's performance in one target bin; sparse target regions exhibit visibly larger dispersion.  
}  
\vspace{-5pt}
\label{fig:monterey-seed-instability}  
\end{wrapfigure}

This confirms that the regions central to DIR are also the least stable under repeated training. Importantly, improved mean performance does not appear to remove this instability. On \textsc{Monterey}, UVote substantially improves bMASE but retains a similar density-dependent variance trend as the vanilla model: the trend starts lower, but still rises as target support decreases. Figure~\ref{fig:monterey-seed-instability} visualizes this effect directly, showing that seed-level variability remains compact in dense target regions but widens in sparse regions. Additionally interesting: many method--dataset pairs exhibit an approximately linear relationship on log--log axes, suggesting an inverse power-law-like scaling
between local target support and seed-level variance. 

We do not propose a new stability remedy here. Instead, we show that stability is a first-order evaluation concern for DIR. If the goal is to improve performance in underrepresented target regions, then reporting only average variance or from a single seed, is insufficient. Tail accuracy and tail stability should be evaluated jointly.

\subsection{Discussion}

The experiments address the three blind spots identified in this paper. \autoref{sec:4_1} establishes \textsc{MuViS-DIR} as a natural extension of the DIR evaluation ecosystem: it provides multimodal time-series tasks with physically meaningful tail regimes, publicly available code, and a benchmark structure suitable for evaluating future DIR methods. \autoref{sec:4_2} further shows that tail performance is improvable with existing DIR methods, but not consistently across the evaluated method set. This positions \textsc{MuViS-DIR} not as a solved application of DIR, but as an open challenge broadening the current data landscape and by this addressing the \textit{Data Blind Spot}. 

\autoref{sec:4_2}  also demonstrates the practical value of bMASE. A single table is sufficient to compare methods across datasets, physical units, and imbalance regimes, while still preserving dataset-level detail. Most importantly, bMASE provides a scalar basis for ranking methods, addressing the ambiguity of many-/median-/few-shot reporting and therefore the \textit{Metric Blind Spot}. 

Finally, \autoref{sec:4_3} directly addresses the \textit{Stability Blind Spot} in the current DIR evaluation ecosystem and confirms that performance instability across seeds in underrepresented regions is not only present but amplified. We do not claim this as a new optimization phenomenon; rather, we show that it has been largely absent from DIR evaluation despite being central to the problem. 

\section{Limitations and Future Work}
\label{sec:limitations}

This study has several limitations. First, while \textsc{MuViS-DIR} broadens DIR evaluation beyond image-centric benchmarks, it is not intended to cover the full space of time-series regression problems. We deliberately focus on virtual sensing as a multimodal, sensor-based sequence-to-value setting because it isolates the transfer of DIR methods to time-series inputs while retaining a single continuous target. Forecasting and sequence-to-sequence prediction introduce additional structure: imbalance may depend not only on the target value, but also on where rare regimes occur along the prediction horizon. Extending the analysis to these settings is therefore an important direction for future work.

Second, our empirical comparison is necessarily selective. We evaluate six representative DIR methods spanning algorithm-level, representation-learning, and architectural approaches, but the DIR literature is growing rapidly. Our results therefore should not be read as a definitive ranking of all DIR methods, but as evidence that existing methods transfer unevenly to multimodal time-series data. Future work should expand the benchmark to newer methods and combinations of orthogonal interventions.

Third, bMAE and bMASE depend on a discretization of the target range. We mitigate this choice by using a shared data-driven binning rule and show in \autoref{app:cont-vs-disc} that the resulting estimates agree closely with a continuous kernel-based alternative. Still, balanced evaluation necessarily requires a choice about local target resolution. Developing equally interpretable continuous alternatives with explicit uncertainty estimates remains an open direction.

Finally, our stability analysis is diagnostic rather than corrective. We show that low-density target regions exhibit higher seed-level variability, but our scope was not to propose a method that explicitly optimizes for stable tail performance. This opens a promising direction for future DIR methods: improving performance in underrepresented regions while also reducing variance across random initialization, data order, and optimization trajectories.

\section{Conclusion}
\label{sec:conclusion}

We revisited DIR from an evaluation perspective and identified three blind spots in the current empirical ecosystem. The first is a \textit{Data Blind Spot}: DIR evaluation remains concentrated in image-based benchmarks, leaving structurally different regression domains underexplored. The second is a \textit{Metric Blind Spot}: many-/medium-/few-shot reporting is useful diagnostically, but does not provide a decision-complete scalar criterion for comparing methods. The third is a \textit{Stability Blind Spot}: the rare target regions that motivate DIR are rarely evaluated for reproducibility across random seeds.

To address these issues, we introduced \textsc{MuViS-DIR}, a multimodal virtual sensing benchmark with continuous time-series regression tasks and practically meaningful tail regimes and we introduced bMASE, a scale-normalized balanced metric that enables cross-method and cross-dataset comparison. Finally, we analyzed DIR methods performance stability as a function of local target support. 

Our experiments show that standard deep learning models can exhibit substantial tail degradation that is masked by global MAE, that existing DIR methods can improve balanced performance but transfer unevenly to multimodal time-series data, and that tail regions are also the least stable across seeds. These findings do not suggest that any single method solves \textsc{MuViS-DIR}. Rather, they show that evaluating DIR requires broader data domains, balanced scalar metrics, and explicit attention to tail reliability.

\bibliographystyle{unsrtnat}
\bibliography{references}

\appendix

\section{Appendix}

This appendix is organized as follows. \autoref{app:math} lists the mathematical notation used throughout the paper. \autoref{app:data_blindspot}, \autoref{app:metric_blindspot}, and \autoref{sec:noise_analysis} provide additional analysis for the data-, metric-, and stability blind spots, respectively. \autoref{app:repro} provides instructions for reproducing the experiments from \autoref{sec:4} and \autoref{app:experiments} contains supplementary results and figures for \autoref{sec:4}. The publicly accessible code for the paper can be found at \url{www.github.com/noah-puetz/muvis-dir}.

\section{Mathematical Notations}
\label{app:math}

\begin{table}[H]
\centering
\caption{Mathematical notation used throughout the paper.}
\label{tab:notation}
\small
\begin{tabularx}{\textwidth}{@{}lX@{}}
\toprule
\textbf{Symbol} & \textbf{Meaning} \\
\midrule
\multicolumn{2}{@{}l}{\textbf{Problem setting}} \\
\midrule
$\mathcal{D}$ & Dataset of input--target pairs. \\
$N$ & Number of samples in a dataset. When referring specifically to the test set for binning, we use $N_{\mathrm{test}}$. \\
$i$ & Sample index, with $i \in \{1,\ldots,N\}$. \\
$\mathcal{X}_i$ & Multimodal sequential input for sample $i$. \\
$x_i^j$ & Time series of modality $j$ for sample $i$. \\
$M$ & Number of input modalities. \\
$j$ & Modality index, with $j \in \{1,\ldots,M\}$. \\
$D^j$ & Number of feature channels in modality $j$. \\
$T^j$ & Number of time steps in modality $j$. \\
$y_i$ & Continuous scalar target value for sample $i$. \\
$y_i(t_0)$ & Target value at reference time $t_0$. \\
$t_0$ & Reference time at which the virtual sensor target is defined. \\
$f_\theta$ & Deep regression model with parameters $\theta$. \\
$\theta$ & Trainable model parameters. \\
$\hat{y}_i$ & Model prediction for sample $i$, i.e. $\hat{y}_i=f_\theta(\mathcal{X}_i)$. \\
$e_i$ & Prediction error for sample $i$, defined as $e_i=y_i-\hat{y}_i$. \\
$p_{\mathrm{train}}(y)$ & Empirical target distribution in the training set. \\
$p_{\mathrm{eval}}(y)$ & Desired evaluation distribution over the target space, typically more balanced than $p_{\mathrm{train}}(y)$. \\
\midrule
\multicolumn{2}{@{}l}{\textbf{Balanced evaluation metrics}} \\
\midrule
$[a,b]$ & Target range used for balanced evaluation. In this work, $a=\min(y_{\mathrm{test}})$ and $b=\max(y_{\mathrm{test}})$. \\
$a$ & Lower endpoint of the test target range. \\
$b$ & Upper endpoint of the test target range. \\
$K$ & Number of equal-width target bins used for bMAE and bMASE. \\
$\mathrm{bin}_k$ & Set of test-sample indices whose target values fall into bin $k$. \\
$k$ & Target-bin index, with $k \in \{1,\ldots,K\}$. \\
$n_k$ & Number of test samples in bin $k$, i.e. $n_k=|\mathrm{bin}_k|$. \\
$y_{\mathrm{train}}$ & Collection of training target values. \\
$y_{\mathrm{test}}$ & Collection of test target values. \\
$c_{\mathrm{ref}}$ & Constant reference predictor used for scaling bMASE; here $c_{\mathrm{ref}}=\mathrm{median}(y_{\mathrm{train}})$. \\
$\mathrm{MAE}$ & Mean absolute error under the empirical test distribution. \\
$\mathrm{bMAE}$ & Balanced mean absolute error, obtained by macro-averaging per-bin MAE across the target range. \\
$\mathrm{MASE}$ & Mean absolute scaled error. \\
$\mathrm{bMASE}$ & Balanced mean absolute scaled error, defined as $\mathrm{bMAE}(f_\theta)/\mathrm{bMAE}(c_{\mathrm{ref}})$. \\
$\mathrm{bMAE}_d(\cdot)$ & bMAE on dataset $d$. \\
$d$ & Dataset index used when aggregating results across datasets. \\
$\mathrm{D}$ & Number of benchmark datasets included in the aggregate score. \\
$\overline{\mathrm{bMASE}}$ & Geometric mean of bMASE values across datasets. \\
\midrule
\multicolumn{2}{@{}l}{\textbf{Binning rule}} \\
\midrule
$w_{\mathrm{FD}}$ & Bin width given by the Freedman--Diaconis rule. \\
$\mathrm{IQR}(y_{\mathrm{test}})$ & Interquartile range of the test targets. \\
$\lceil \cdot \rceil$ & Ceiling operator. \\
\midrule
\multicolumn{2}{@{}l}{\textbf{Stability analysis}} \\
\midrule
$s$ & Random seed index. \\
$S$ & Number of random seeds. In the experiments, $S=10$. \\
$\sigma_k$ & Standard deviation of per-bin MAE across seeds for bin $k$. \\
$\rho_{\mathrm{Spearman}}$ & Spearman rank correlation between bin sample count and seed-level per-bin instability. \\
\bottomrule
\end{tabularx}
\end{table}

\section{Data Blind Spot Appendix}
\label{app:data_blindspot}

This appendix supplements the data-blind-spot discussion in
\autoref{sec:1} along two complementary axes. \autoref{app:data_overview} provides a comprehensive snapshot of the DIR methodological landscape and architectures they were originally evaluated on, and indicating which combinations come with public training code. The purpose is to make explicit how narrow the empirical evidence behind current DIR claims actually is, and why a domain-extended benchmark such as \textsc{MuViS} is needed. \autoref{app:muvis} then characterises \textsc{MuViS} itself, both statistically (target imbalance, effective support of the training distribution) and operationally (why uniform performance across the target range is a domain requirement, not just a statistical preference).

\subsection{DIR Methods and Dataset Landscape}
\label{app:data_overview}

\autoref{tab:dir_methods_summary} makes the data blind spot quantitatively visible: 18 of the 19 surveyed methods include AgeDB-DIR or IMDB-WIKI-DIR as their primary evaluation, and only a small subset of works ventures beyond facial-image regression.

{\small
\setlength{\tabcolsep}{4pt}
\begin{longtable}{@{}p{0.9cm} p{2.2cm} p{1.9cm} p{3.0cm} p{2.7cm} p{0.6cm} p{0.85cm}@{}}
\caption{Overview of DIR methods, datasets, and backbone architectures.
Datasets and models in \textbf{bold} indicate that public training
code is available for the corresponding combination.
``Algo.''\ stands for algorithm-level methods, ``Repr.''\ for
representation-learning methods, and ``Archit.''\ for architectural
methods. The \emph{Orth.}\ column indicates whether the method is
orthogonal to the backbone architecture and can be combined with
other methods.}
\label{tab:dir_methods_summary}\\
\toprule
\textbf{Type} & \textbf{Method} & \textbf{Venue} & \textbf{Datasets} & \textbf{Models} & \textbf{Orth.} & \textbf{Repo} \\
\midrule
\endfirsthead

\multicolumn{7}{l}{\textit{Table~\ref{tab:dir_methods_summary} -- continued from previous page}} \\
\toprule
\textbf{Type} & \textbf{Method} & \textbf{Venue} & \textbf{Datasets} & \textbf{Models} & \textbf{Orth.} & \textbf{Repo} \\
\midrule
\endhead

\midrule
\multicolumn{7}{r}{\textit{Continued on next page}} \\
\endfoot

\bottomrule
\endlastfoot

Algo. & LDS~\citep{yang_delving_2021} & ICML~2021 & \textbf{IMDB-WIKI-DIR}\newline \textbf{AgeDB-DIR}\newline \textbf{STS-B-DIR}\newline \textbf{NYUD2-DIR}\newline SHHS-DIR & \textbf{ResNet-50}\newline \textbf{BiLSTM+GloVe}\newline CNN-RNN & \cmark & \href{https://github.com/YyzHarry/imbalanced-regression}{GitHub} \\
 & Balanced MSE~\citep{ren_balanced_2022} & CVPR~2022 & \textbf{IMDB-WIKI-DIR}\newline \textbf{NYUD2-DIR}\newline IHMR & \textbf{ResNet-50}\newline SPIN & \cmark & \href{https://github.com/jiawei-ren/BalancedMSE}{GitHub} \\
 & VIR~\citep{wang_variational_2023} & NeurIPS~2023 & \textbf{IMDB-WIKI-DIR}\newline \textbf{AgeDB-DIR}\newline STS-B-DIR\newline NYUD2-DIR & \textbf{ResNet-50}\newline BiLSTM+GloVe & \xmark & \href{https://github.com/Wang-ML-Lab/variational-imbalanced-regression}{GitHub} \\
 & DenseLoss~\citep{steininger_density-based_2021} & \textit{Mach.\ Learn.}~2021 & Synthetic Data & MLP & \cmark & \href{https://github.com/SteiMi/denseweight}{GitHub} \\
 & Focal-R~\citep{yang_delving_2021, lin_focal_2018} & ICML~2021 & \textbf{IMDB-WIKI-DIR}\newline \textbf{AgeDB-DIR}\newline \textbf{STS-B-DIR}\newline \textbf{NYUD2-DIR}\newline SHHS-DIR & \textbf{ResNet-50}\newline \textbf{BiLSTM+GloVe}\newline CNN-RNN & \cmark & \href{https://github.com/YyzHarry/imbalanced-regression}{GitHub} \\
 & INV~\& SQINV~\citep{yang_delving_2021} & ICML~2021 & \textbf{IMDB-WIKI-DIR}\newline \textbf{AgeDB-DIR}\newline \textbf{STS-B-DIR}\newline \textbf{NYUD2-DIR}\newline SHHS-DIR & \textbf{ResNet-50}\newline \textbf{BiLSTM+GloVe}\newline CNN-RNN & \cmark & \href{https://github.com/YyzHarry/imbalanced-regression}{GitHub} \\
 & RRT~\citep{yang_delving_2021} & ICML~2021 & \textbf{IMDB-WIKI-DIR}\newline \textbf{AgeDB-DIR}\newline \textbf{STS-B-DIR}\newline \textbf{NYUD2-DIR}\newline SHHS-DIR & \textbf{ResNet-50}\newline \textbf{BiLSTM+GloVe}\newline CNN-RNN & \cmark & \href{https://github.com/YyzHarry/imbalanced-regression}{GitHub} \\
 & Dist Loss~\citep{nie_dist_2025} & ICLR~2025 & \textbf{IMDB-WIKI-DIR}\newline \textbf{AgeDB-DIR}\newline ECG-K-DIR & \textbf{ResNet-50}\newline Net1D & \cmark & \href{https://github.com/Ngk03/DIR-Dist-Loss}{GitHub} \\
\midrule

Repr. & FDS~\citep{yang_delving_2021} & ICML~2021 & \textbf{IMDB-WIKI-DIR}\newline \textbf{AgeDB-DIR}\newline \textbf{STS-B-DIR}\newline \textbf{NYUD2-DIR}\newline SHHS-DIR & \textbf{ResNet-50}\newline \textbf{BiLSTM+GloVe}\newline CNN-RNN & \cmark & \href{https://github.com/YyzHarry/imbalanced-regression}{GitHub} \\
 & RankSim~\citep{gong_ranksim_2022} & ICML~2022 & \textbf{IMDB-WIKI-DIR}\newline \textbf{AgeDB-DIR}\newline \textbf{STS-B-DIR} & \textbf{ResNet-50}\newline \textbf{BiLSTM+GloVe} & \cmark & \href{https://github.com/BorealisAI/ranksim-imbalanced-regression}{GitHub} \\
 & ConR~\citep{keramati_conr_2023} & ICLR~2024 & \textbf{IMDB-WIKI-DIR}\newline \textbf{AgeDB-DIR}\newline \textbf{NYUD2-DIR}\newline MPIIGaze-DIR & \textbf{ResNet-50}\newline LeNet & \cmark & \href{https://github.com/BorealisAI/ConR/tree/main}{GitHub} \\
 & RnC~\citep{zha_rank-n-contrast_2023} & NeurIPS~2023 & IMDB-WIKI-DIR\newline \textbf{AgeDB-DIR}\newline TUAB\newline MPIIGaze-DIR\newline SkyFinder & \textbf{ResNet-18}\newline \textbf{ResNet-50} & \xmark\ (\cmark) & \href{https://github.com/kaiwenzha/Rank-N-Contrast/tree/main}{GitHub} \\
 & Geom.\ Rep.~\citep{dong_improve_2025} & CVPR~2025 & UCI-DIR\newline AgeDB-DIR\newline IMDB-WIKI-DIR\newline \textbf{STS-B-DIR} & ResNet-50\newline \textbf{BiLSTM+GloVe}\newline MLP & \cmark & \href{https://github.com/yilei-wu/imbalanced-regression}{GitHub} \\
 & Ordinal Ent.~\citep{zhang_improving_2023} & ICLR~2023 & \textbf{Synthetic data}\newline NYU-Depth-v2\newline AgeDB-DIR\newline SHTech & \textbf{MLP}\newline NeW-CRFs\newline ResNet-50\newline DeepONet\newline CSRNet & \cmark & \href{https://github.com/needylove/OrdinalEntropy/tree/main}{GitHub} \\
 & PRIME~\citep{lim_prime_2025} & ICML~2025 & AgeDB-DIR\newline IMDB-WIKI-DIR\newline NYUD2-DIR\newline STS-B-DIR & ResNet-50\newline BiLSTM+GloVe & \cmark & --- \\
\midrule

Archit. & UVOTE~\citep{jiang_uncertainty_2024} & GCPR~2024 & \textbf{IMDB-WIKI-DIR}\newline \textbf{AgeDB-DIR}\newline Wind\newline \textbf{STS-B-DIR} & \textbf{ResNet-50}\newline \textbf{ResNet-18}\newline \textbf{BiLSTM} & \xmark & \href{https://github.com/SherryJYC/UVOTE/tree/main}{GitHub} \\
 & HCA~\citep{xiong_deep_2024} & CVPR~2024 & IMDB-WIKI-DIR\newline AgeDB-DIR\newline NYUDv2-DIR\newline SHTech & ResNet-50\newline VGG16 & \xmark & \href{https://github.com/xhp-hust-2018-2011/HCA}{GitHub} \\
 & Multi-Class.~\citep{lin_let_2024} & ICANN~2024 & IMDB-WIKI-DIR\newline AgeDB-DIR\newline STS-B-DIR & ResNet-50\newline BiLSTM+GloVe & \xmark & --- \\
 & Im-Context~\citep{nejjar_im-context_2024} & TMLR~2024 & IMDB-WIKI-DIR\newline AgeDB-DIR\newline STS-B-DIR\newline \textbf{Boston\newline Concrete\newline Abalone\newline Communities\newline Kin8nm\newline Airfoil} & GPT2\newline \textbf{PFN} & \xmark & \href{https://github.com/ismailnejjar/IM-Context}{GitHub} \\

\end{longtable}
}

\subsection{\textsc{MuViS} Dataset Overview}
\label{app:muvis}

We characterise the \textsc{MuViS} datasets along two complementary
dimensions. The \emph{imbalance ratio}~(IR) summarises the contrast
between dense head and sparse tails of the training target
distribution, while the \emph{effective support}~$S_{\mathrm{eff}}$
quantifies how much of the available target range the training
distribution actually uses. Together, the two metrics distinguish
distributions that are similarly skewed but differ in coverage, and
are reported alongside conventional moments
(skew, excess kurtosis) in \autoref{tab:dataset_stats}.
\autoref{fig:target_distributions} additionally visualises all nine
target distributions.

\paragraph{Imbalance ratio.}
\label{app:IR}
The imbalance ratio is defined as the ratio of sample density in the
head of the training distribution to sample density in its tails:
\begin{equation}
  \mathrm{IR}
  \;=\;\frac{d_{\mathrm{head}}}{d_{\mathrm{tail}}}\,,
\end{equation}
where, 
\begin{equation}
  d_{\mathrm{head}}
  \;=\; \frac{\bigl|\{i : Q_{25} \le y_i \le Q_{75}\}\bigr|}
              {Q_{75}-Q_{25}}\,,
              \qquad
  d_{\mathrm{tail}}
  \;=\; \frac{\bigl|\{i : y_i < Q_{10}\}\bigr|
              + \bigl|\{i : y_i > Q_{90}\}\bigr|}
              {(Q_{10}-y_{\min}) + (y_{\max}-Q_{90})}\,,
\end{equation}
where $Q_{p}$ denotes the $p$-th percentile of the training targets.
The quantile thresholds determine \emph{which} samples count as head
or tail, by construction the head always contains $50\%$ of training
samples and the two tails together $20\%$, while the densities
$d_{\mathrm{head}}$ and $d_{\mathrm{tail}}$ are measured in
target-space units, matching the equal-width binning that
$\mathrm{bMAE}$ and $\mathrm{bMASE}$ apply at evaluation time. A
uniform target distribution yields $\mathrm{IR}=1$. For sharply
peaked distributions the $20\%$ of tail samples are spread over a
wider target range, driving $d_{\mathrm{tail}}$ down and IR up.

\paragraph{Effective support.}
\label{app:S_eff}
While IR captures the head-to-tail contrast, two distributions with
identical IR can still differ substantially in how much of the target
range they cover (e.g.\ a heavily skewed distribution versus a
bimodal one). The effective support $S_{\mathrm{eff}}$ provides a
single global measure of this coverage. We fit a Gaussian KDE with
Silverman bandwidth to the training targets, evaluate it on a grid
of $G$ equally spaced points spanning the target range, normalise to
a discrete distribution~$q$, and define:
\begin{equation}
  S_{\mathrm{eff}}
  \;=\; \frac{\exp\bigl(H(q)\bigr)}{G}\,,
  \qquad
  H(q) \;=\; -\sum_{g=1}^{G} q_g \log q_g\,,
  \label{eq:seff}
\end{equation}
where $H(q)$ is the Shannon entropy of $q$. Intuitively,
$S_{\mathrm{eff}}$ answers the question \emph{what fraction of the
target range is effectively occupied by the training distribution?}
A value of $1$ corresponds to a perfectly uniform distribution; lower
values indicate that the mass is concentrated in a smaller fraction
of the range. $S_{\mathrm{eff}}$ is the continuous analogue of the
``effective number of classes'' notion from imbalanced classification.
Because the smoothing is controlled by the KDE bandwidth rather than
by~$G$, $S_{\mathrm{eff}}$ is stable across discretisation choices
($<\!0.6\%$ variation over a $20\!\times$ range of~$G$).

\paragraph{Dataset statistics.}
\autoref{tab:dataset_stats} reports both metrics together with sample
sizes, sequence length~$T$, input channels~$D$, and conventional
distributional moments. The nine \textsc{MuViS} tasks span more than
an order of magnitude in IR (from~2.0 on Battery~SoC to~41.2 on
Monterey) and a corresponding range in $S_{\mathrm{eff}}$ (from~$0.92$
down to~$0.25$). This range is a deliberate property of the benchmark:
it allows DIR methods to be tested across the full spectrum from
near-uniform to extremely peaked target distributions, rather than
within the narrow band covered by the predominantly Gaussian
facial-age benchmarks.

\begin{table}[ht]
\caption{\textsc{MuViS} dataset statistics and imbalance characterisation.
$T$: sequence length; $D$: input channels;
IR: imbalance ratio ($\uparrow$\,= more imbalanced);
$S_{\mathrm{eff}}$: effective support ($\downarrow$\,= more concentrated).
Tasks ordered by decreasing IR.}
\label{tab:dataset_stats}
\centering
\small
\begin{tabular}{lrrrrrrrr}
\toprule
Dataset & $n_{\mathrm{train}}$ & $n_{\mathrm{test}}$ & $T$ & $D$
        & Skew & Kurt. & IR\,$\uparrow$ & $S_{\mathrm{eff}}$\,$\downarrow$ \\
\midrule
Monterey                & 120{,}777 & 21{,}357  & 20  & 22 & $+0.20$ & $\phantom{-}3.27$  & 41.2 & 0.25 \\
Targa~'13               & 33{,}520  & 11{,}109  & 20  & 22 & $+0.35$ & $\phantom{-}1.96$  & 26.0 & 0.32 \\
Targa~'14               & 46{,}739  & 9{,}036   & 20  & 22 & $-0.18$ & $\phantom{-}0.16$  & 16.8 & 0.43 \\
PM\textsubscript{2.5}   & 11{,}918  & 5{,}048   & 24  & 9  & $+2.09$ & $\phantom{-}6.86$  & 15.6 & 0.42 \\
PM\textsubscript{10}    & 11{,}918  & 5{,}048   & 24  & 9  & $+2.21$ & $\phantom{-}6.87$  & 12.9 & 0.44 \\
Chem.\ Conc.            & 240{,}500 & 470{,}500 & 20  & 33 & $-0.01$ & $-0.05$            & 12.1 & 0.45 \\
Tire Temp.              & 1{,}384   & 280       & 50  & 11 & $+0.18$ & $-0.60$            & \phantom{0}8.9  & 0.55 \\
Heart Rate              & 51{,}757  & 12{,}940  & 512 & 6  & $+0.90$ & $\phantom{-}0.85$  & \phantom{0}8.0  & 0.60 \\
Battery SoC             & 199{,}827 & 158{,}126 & 120 & 7  & $-0.20$ & $-1.01$            & \phantom{0}2.0  & 0.92 \\
\bottomrule
\end{tabular}
\end{table}

\subsubsection{Domain motivation for tail-region reliability}

The statistical imbalance summarised above is necessary but not sufficient to motivate balanced evaluation. What makes \textsc{MuViS} a meaningful DIR benchmark is that, for every task, the distributional tails correspond to operationally critical regimes where predictive failures carry disproportionate consequences.

\begin{enumerate}
  \item \textbf{PM\textsubscript{2.5}\,/\,PM\textsubscript{10}.} The Beijing Multi-Site Air Quality data~\citep{zhang_cautionary_2017} pose a virtual sensing task in which the target is particulate-matter concentration and the inputs combine pollutant and meteorological measurements. Extreme concentrations exceed air-quality-index thresholds that trigger public-health advisories and emergency traffic restrictions~\citep{chen_evidence_2013}.

  \item \textbf{Monterey\,/\,Targa~'13\,/\,Targa~'14.} The Revs Program Vehicle Dynamics Database integrates multimodal sensing sources (driver inputs, wheel and chassis measurements, GNSS-aided inertial navigation). Following \citet{brandt_faults_2025}, we define a virtual sensing task of estimating the vehicle's lateral velocity~$v_y$, a key state for stability assessment and control that is typically not directly available from low-cost on-board sensing. Extreme values correspond to near-limit handling regimes marking the onset of tire-grip loss and vehicle instability. These are precisely the safety-critical conditions that electronic stability control systems are designed to navigate~\citep{kegelman_insights_2017}.

  \item \textbf{Tire Temp.} \textsc{MuViS} additionally includes a high-performance autonomous-driving dataset that records vehicle state from RTK-GPS alongside control inputs, actuator states, and the target real-time tire temperature~\citep{mori_vehicle_2025}. Extreme surface temperatures indicate accelerated tread degradation and elevated blowout risk; accurate tail prediction is therefore essential for high-performance autonomous applications~\citep{brandt_muvis_2026}.

  \item \textbf{Chem.\ Conc.} Tails flag abnormal operating states corresponding to the 21 benchmarked fault scenarios of the Tennessee Eastman process \citep{downs_plant-wide_1993}. A model that fails to predict these extremes effectively fails to detect process faults.

  \item \textbf{Battery SoC.} Extremes represent overcharge and deep-discharge regimes where cell degradation accelerates and safety margins narrow. Prediction bias at these boundaries can trigger premature or delayed charge termination \citep{kollmeyer_panasonic_2018, mondal_estimating_2024}.

  \item \textbf{Heart Rate.} Extremes indicate exercise-induced tachycardia or arrhythmia events that are clinically significant yet inherently rare in ambulatory recordings~\citep{reiss_deep_2019}.
\end{enumerate}

In each case, an evaluation protocol that down-weights tail performance in proportion to its empirical frequency, as standard MAE and~$R^{2}$ implicitly do, would systematically reward models that fail in exactly the regimes the underlying application cares most about.

\section{Metric Blind Spot Appendix}
\label{app:metric_blindspot}

This appendix provides the empirical analysis underlying the metric choices in \autoref{sec:3_2}. We study (i)~the distribution invariance of standard and balanced regression metrics, and (ii)~the discrete-versus-continuous estimation question for the bMAE numerator that underlies bMASE.

\subsection{Distribution invariance of standard and balanced metrics}

In this section we analyse the sensitivity of different evaluation metrics to the test distribution and show why several widely used metrics are not aligned with the evaluation objective of DIR. \citet{puetz_deconstructing_2026} describes the fundamental challenge in DIR as a ``mismatch between the training objective and the evaluation goal'': the very act of addressing imbalance implies that we value uniform performance across all target regions more than their representation in the training data would suggest, which is equivalent to a distribution shift in the target space and therefore breaks the standard i.i.d.\ assumption~\citep{puetz_deconstructing_2026, shalev-shwartz_understanding_2014, quionero-candela_dataset_2009}. Building on this line of argument, our goal is a single scalar metric that ranks models consistently regardless of the test-set composition. The experiment below shows how a metric's sensitivity to the test distribution can shift both model rankings and performance estimates, and demonstrates that bMAE and bMASE are robust against this effect. 

\subsubsection{Experimental setup}\label{app:metric-setup}

To isolate metric properties from model properties, we construct four \emph{deterministic} synthetic models whose error profile is a fixed function of the target value. Given a test sample with true target~$y$, we define
$d = |y - \mu_{\mathrm{train}}|/(0.4\,(b-a))$ and
$s = \mathrm{sign}(y - \mu_{\mathrm{train}})$, and study:
\begin{itemize}
  \item \textbf{A} (tail collapse):      $\hat{y} = y + (0.5 + 10\,d^{2})\,s$
  \item \textbf{B} (uniformly mediocre): $\hat{y} = y + (2.5 + 0.3\,d^{2})\,s$
  \item \textbf{C} (uniformly good):     $\hat{y} = y + (1.0 + 0.3\,d^{2})\,s$
  \item \textbf{D} (tail specialist):    $\hat{y} = y + (0.5 + 10\,e^{-3d})\,s$
\end{itemize}
The four models cover the failure modes most relevant to DIR: a typical empirical-risk-minimisation profile that collapses in the tails (A), a uniformly mediocre baseline (B), a uniformly strong predictor (C), and a tail-specialist that sacrifices head accuracy in exchange for better tail behaviour (D). Because the errors are fully deterministic functions of~$y$, any variation in a metric across test conditions is attributable purely to the metric's sensitivity to the test-set distribution rather than to randomness in the predictions.

\begin{figure}
  \centering{
  \includegraphics[width=\linewidth]{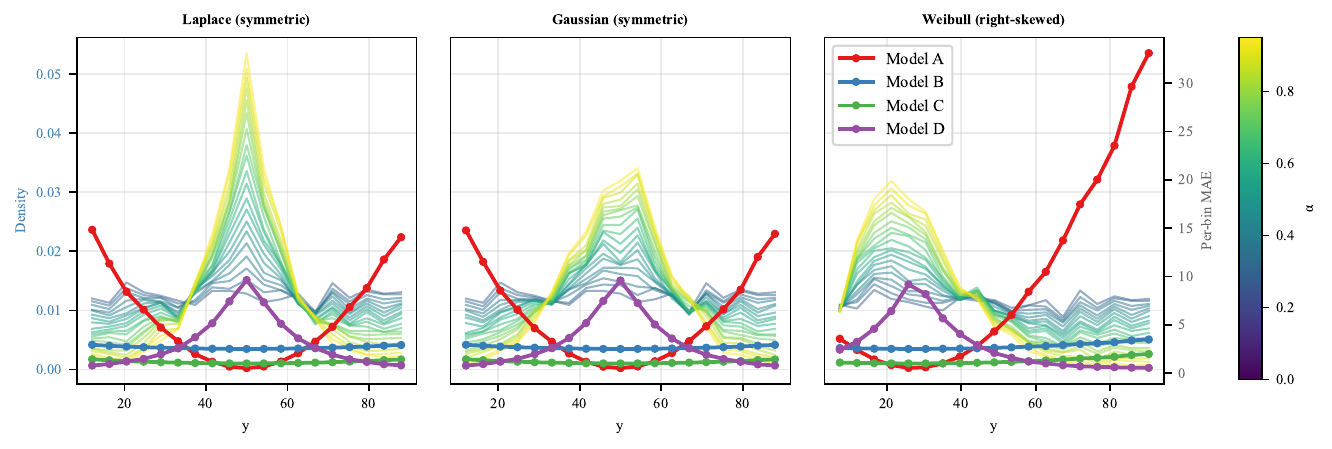}
  \caption{Experimental Setup wiht all three distributions and the corresponding error profiles of the four synthetic models overlaid}
  \label{fig:experiment_setup}}
\end{figure}

For each experiment, test sets are parametrised by $\alpha\in[0, 0.95]$: at $\alpha=0$ the test set is drawn uniformly over the target range, while at $\alpha=0.95$ it concentrates around the corresponding distribution. We test three target distribution families: Gaussian (symmetric), Laplace (symmetric, heavy-tailed), and Weibull (right-skewed). The cap at $\alpha=0.95$ guarantees a minimum of $5\%$ uniform samples, which prevents empty edge bins and avoids the discontinuities that occur when a test set has zero support in parts of the target range. All three distributions and the corresponding error profiles can be seen in \autoref{fig:experiment_setup}.

For each metric we report (i)~the coefficient of variation $\mathrm{CV}=\mathrm{std}/|\mathrm{mean}|$ across the $\alpha$~sweep (lower $=$ more stable) and (ii)~ranking stability: the number of distinct model orderings observed and Kendall's~$\tau$ between the rankings at $\alpha=0$ and $\alpha=0.95$.

\subsubsection{Results}
\label{app:dist-invariance}

\begin{figure}
  \centering
  \includegraphics[width=0.8\linewidth]{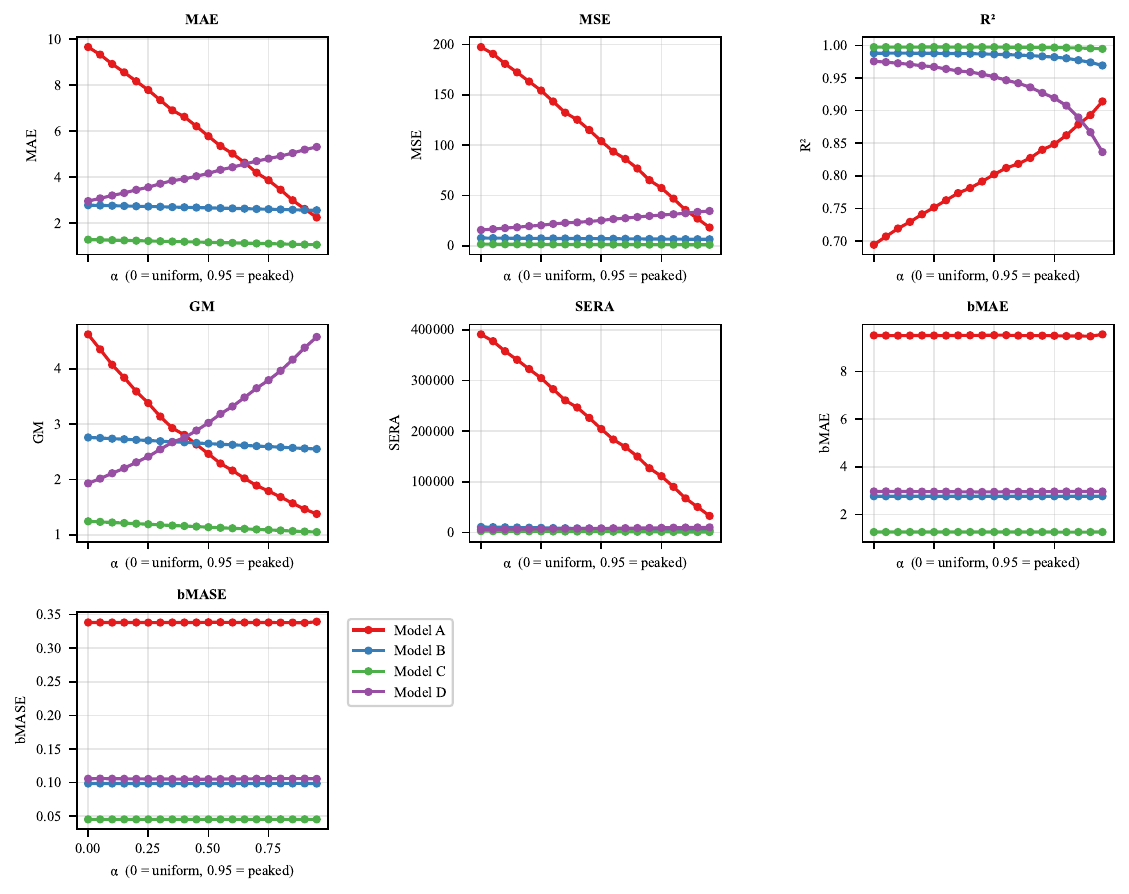}
  \caption{Metric values across test-distribution shifts
    ($\alpha=0$: uniform; $\alpha=0.95$: Weibull-peaked) for the four deterministic models. Standard metrics (top) and SERA drift substantially with~$\alpha$, while the balanced metrics (bMAE,bMASE) remain flat and produce a single stable ranking.}
  \label{fig:grid-weibull}
\end{figure}

\autoref{fig:grid-weibull} shows the values of seven candidate metrics across the $\alpha$~sweep on the Weibull distribution the most challenging setting due to its heavy right skew. Standard MAE, MSE, $R^{2}$, the geometric mean of absolute errors (GM;~\citet{yang_delving_2021}), and SERA~\citep{ribeiro_imbalanced_2020} all drift with~$\alpha$ for at least one model, indicating that the reported metric value depends on the empirical test-set composition rather than only on the model itself. The balanced metrics (bMAE and bMASE) remain flat across~$\alpha$.

\begin{figure}
  \centering
  \includegraphics[width=0.8\linewidth]{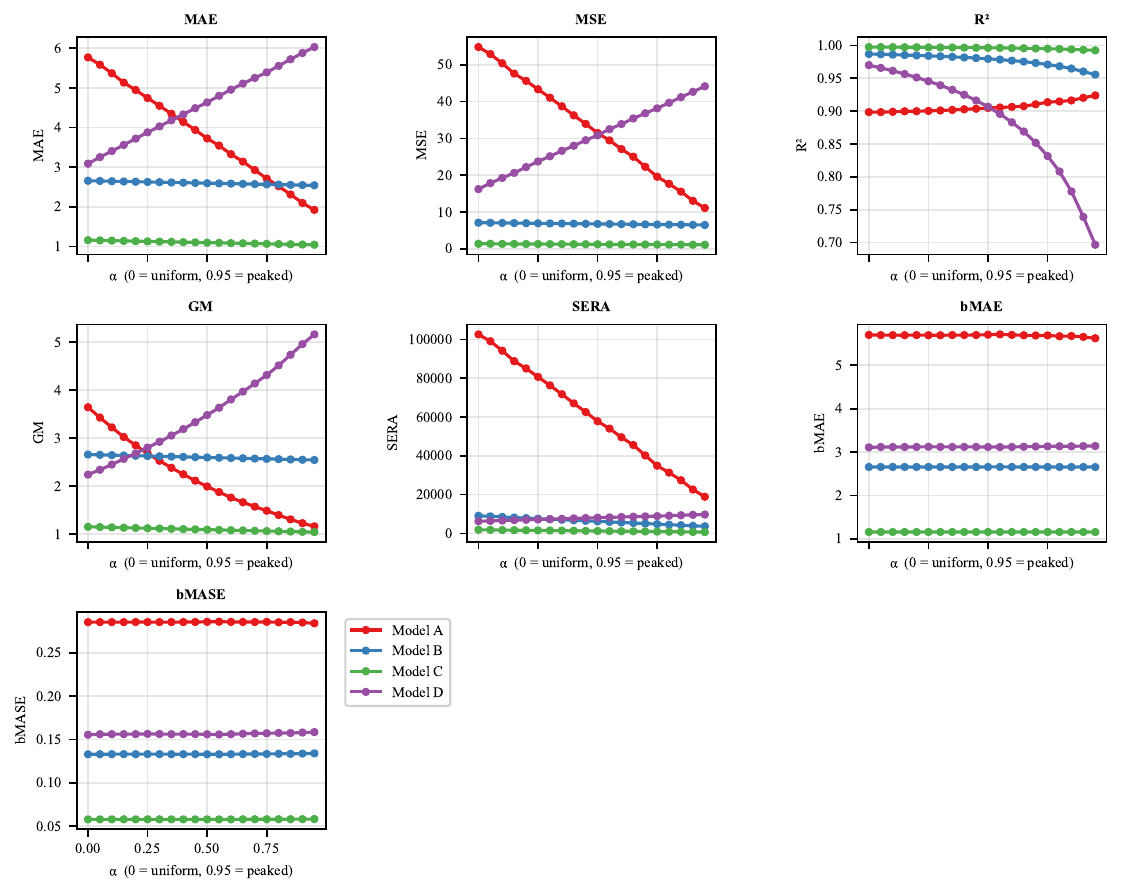}\\[2pt]
  \includegraphics[width=0.8\linewidth]{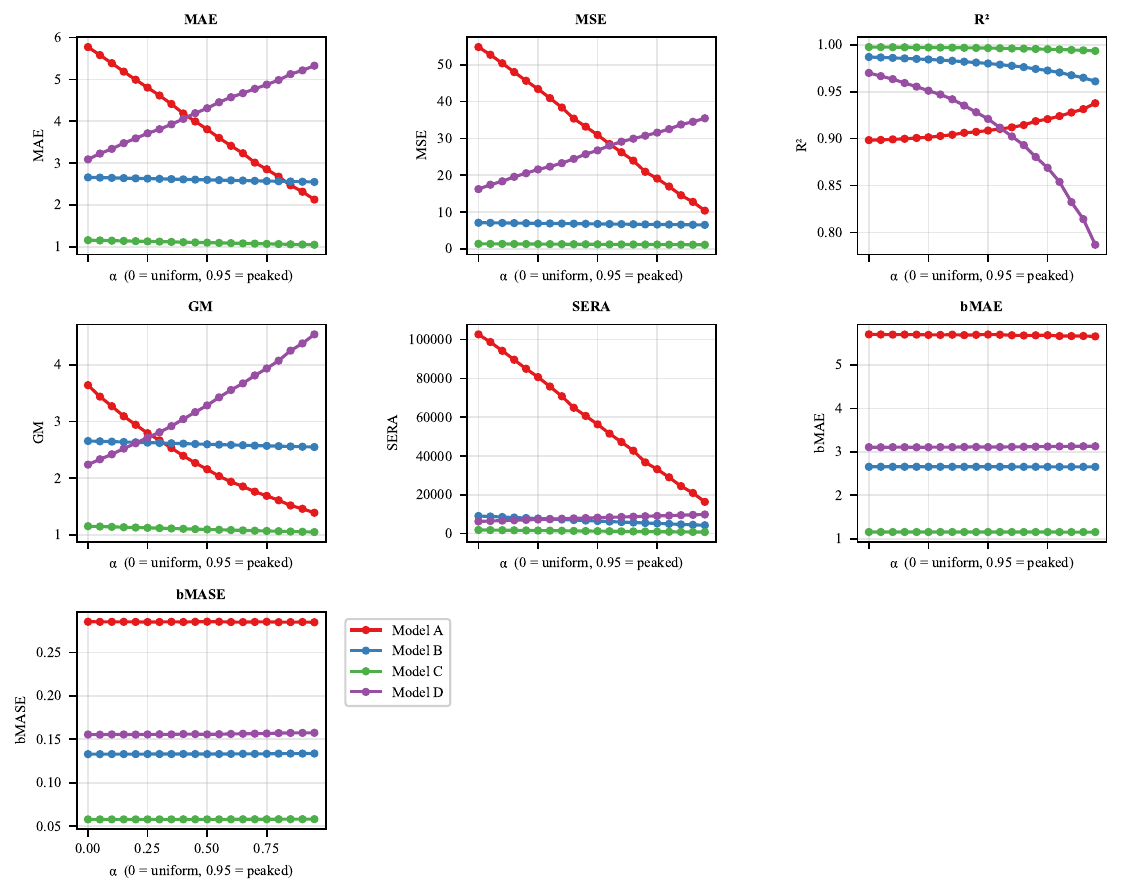}
  \caption{Same analysis as Figure~\ref{fig:grid-weibull} for the
    Laplace (top) and Gaussian (bottom) target distributions. The
    stability pattern of the balanced metrics is consistent across
    distribution families.}
  \label{fig:grid-gauss-laplace}
\end{figure}

\autoref{tab:cv-comparison} quantifies these observations across all three distribution families by computing the per-model CV across the $\alpha$~sweep and the resulting ranking stability. SERA, despite its threshold-free design, is among the least stable metrics (per-model CV up to $0.515$ on Weibull and $0.452$ on Gaussian). This arises because SERA computes an unnormalised \emph{sum} of squared errors at each relevance threshold; when the test distribution shifts, the sample count above each threshold changes, conflating test-set composition with model quality. The proposed balanced metrics achieve near-perfect invariance: per-model CV~$\le 0.005$ across all distributions and exactly one unique ranking across all~$\alpha$ values, with $\tau(0\!\leftrightarrow\!0.95)=+1.00$. \autoref{fig:grid-gauss-laplace} additionally visualises the $\alpha$-sweep curves for Laplace and Gaussian targets to confirm that the qualitative pattern of \autoref{fig:grid-weibull} holds in the symmetric setting.

We emphasise that the analysis in \autoref{fig:grid-weibull}, \autoref{fig:grid-gauss-laplace}, and \autoref{tab:cv-comparison} is not intended as a blanket judgement against MAE, MSE, $R^{2}$, GM, or SERA. Each of these metrics has well-established uses and remains appropriate in many regression settings. What we report is specifically the behaviour that becomes problematic under the \emph{uniform-performance objective of DIR}: when the goal is to characterise model quality consistently across the entire target range, sensitivity to test-set composition is undesirable, and our experiments show that the balanced metrics are the only candidates that meet this criterion.

\begin{table}
\centering
\caption{Distribution invariance of evaluation metrics across the three target distribution families. We report the per-model CV across the $\alpha$~sweep (lower~$=$~more stable), the number of distinct model rankings observed across $\alpha\in[0,0.95]$, and Kendall's~$\tau$ between the rankings at the endpoints. The proposed balanced metrics (bMAE, bMASE) maintain near-perfect invariance across all three settings.}
\label{tab:cv-comparison}
\small
\setlength{\tabcolsep}{4pt}
\begin{tabular}{llccccccc}
\toprule
 & & \multicolumn{4}{c}{CV across $\alpha$ $\downarrow$}
 & \multirow{2}{*}{\#\,Unique $\downarrow$}
 & \multirow{2}{*}{$\tau(0\!\leftrightarrow\!0.95)$ $\uparrow$} \\
\cmidrule(lr){3-6}
Distribution & Metric & A & B & C & D & & \\
\midrule
\multirow{7}{*}{Gaussian}
 & MAE        & 0.285 & 0.013 & 0.161 & 0.030 & 3 & $+0.60$ \\
 & MSE        & 0.424 & 0.026 & 0.225 & 0.064 & 2 & $+0.80$ \\
 & $R^{2}$    & 0.013 & 0.008 & 0.058 & 0.001 & 2 & $+0.80$ \\
 & GM         & 0.293 & 0.013 & 0.213 & 0.029 & 4 & $+0.40$ \\
 & SERA       & 0.452 & 0.221 & 0.133 & 0.254 & 2 & $+0.80$ \\
\cmidrule(lr){2-8}
 & bMAE       & \textbf{0.002} & \textbf{0.000} & \textbf{0.002} & \textbf{0.000} & \textbf{1} & \textbf{$+1.00$} \\
 & bMASE      & \textbf{0.001} & \textbf{0.002} & \textbf{0.004} & \textbf{0.002} & \textbf{1} & \textbf{$+1.00$} \\
\midrule
\multirow{7}{*}{Laplace}
 & MAE        & 0.305 & 0.014 & 0.195 & 0.032 & 3 & $+0.60$ \\
 & MSE        & 0.408 & 0.028 & 0.279 & 0.066 & 2 & $+0.80$ \\
 & $R^{2}$    & 0.008 & 0.009 & 0.087 & 0.001 & 2 & $+0.80$ \\
 & GM         & 0.343 & 0.013 & 0.251 & 0.031 & 4 & $+0.40$ \\
 & SERA       & 0.425 & 0.259 & 0.130 & 0.287 & 2 & $+0.80$ \\
\cmidrule(lr){2-8}
 & bMAE       & \textbf{0.003} & \textbf{0.000} & \textbf{0.003} & \textbf{0.001} & \textbf{1} & \textbf{$+1.00$} \\
 & bMASE      & \textbf{0.001} & \textbf{0.002} & \textbf{0.005} & \textbf{0.002} & \textbf{1} & \textbf{$+1.00$} \\
\midrule
\multirow{7}{*}{Weibull}
 & MAE        & 0.379 & 0.026 & 0.173 & 0.058 & 3 & $+0.60$ \\
 & MSE        & 0.506 & 0.054 & 0.229 & 0.129 & 2 & $+0.80$ \\
 & $R^{2}$    & 0.078 & 0.005 & 0.040 & 0.001 & 2 & $+0.80$ \\
 & GM         & 0.363 & 0.024 & 0.260 & 0.052 & 3 & $+0.40$ \\
 & SERA       & 0.515 & 0.270 & 0.153 & 0.333 & 2 & $+0.80$ \\
\cmidrule(lr){2-8}
 & bMAE       & \textbf{0.002} & \textbf{0.000} & \textbf{0.002} & \textbf{0.000} & \textbf{1} & \textbf{$+1.00$} \\
 & bMASE      & \textbf{0.001} & \textbf{0.001} & \textbf{0.003} & \textbf{0.001} & \textbf{1} & \textbf{$+1.00$} \\
\bottomrule
\end{tabular}
\end{table}

\subsection{Theoretical grounding and continuous estimation of bMAE}
\label{app:cont-vs-disc}

The bMAE used in both \autoref{eq:bmae} and  in bMASE (\autoref{eq:bmase}) is a discrete estimator of the conditional absolute error integrated under the uniform measure over the target range:
\begin{equation}\label{eq:bmae-integral}
  \mathrm{bMAE}(f_{\theta})
  \;\approx\;
  \frac{1}{b-a}\int_{a}^{b}
  \mathbb{E}\!\bigl[|Y-\hat{Y}|\,\bigm|\,Y\!=\!y\bigr]\,\mathrm{d}y\,.
\end{equation}
The uniform measure $\frac{1}{b-a}\,\mathrm{d}y$ ensures that every point in the target range contributes equally to the integral, independent of the empirical test-set density. The binned estimator in \autoref{eq:bmae} approximates this integral via the midpoint quadrature rule, with each bin's per-sample MAE serving as the function value at that bin's centre.

A continuous alternative is to estimate the integrand with Nadaraya--Watson (NW) kernel regression and integrate over a uniform grid:
\begin{equation}\label{eq:nw}
  \widehat{\mathrm{bMAE}}_{\mathrm{NW}}
  = \frac{1}{G}\sum_{g=1}^{G}
  \frac{
    \sum_{i=1}^{N} K_{h}(y_{g}-y_{i})\,|y_{i}-\hat{y}_{i}|
  }{
    \sum_{i=1}^{N} K_{h}(y_{g}-y_{i})
  }\,,
\end{equation}
where $K_{h}$ is a Gaussian kernel with bandwidth~$h$ chosen by Silverman's rule~\citep{silverman_density_2018} and $G\!=\!500$ uniformly spaced grid points. The grid density~$G$ is a numerical integration parameter rather than a statistical one; results are stable from $G\!\approx\!100$ onward.

\begin{figure}
  \centering
  \includegraphics[width=0.8\linewidth]{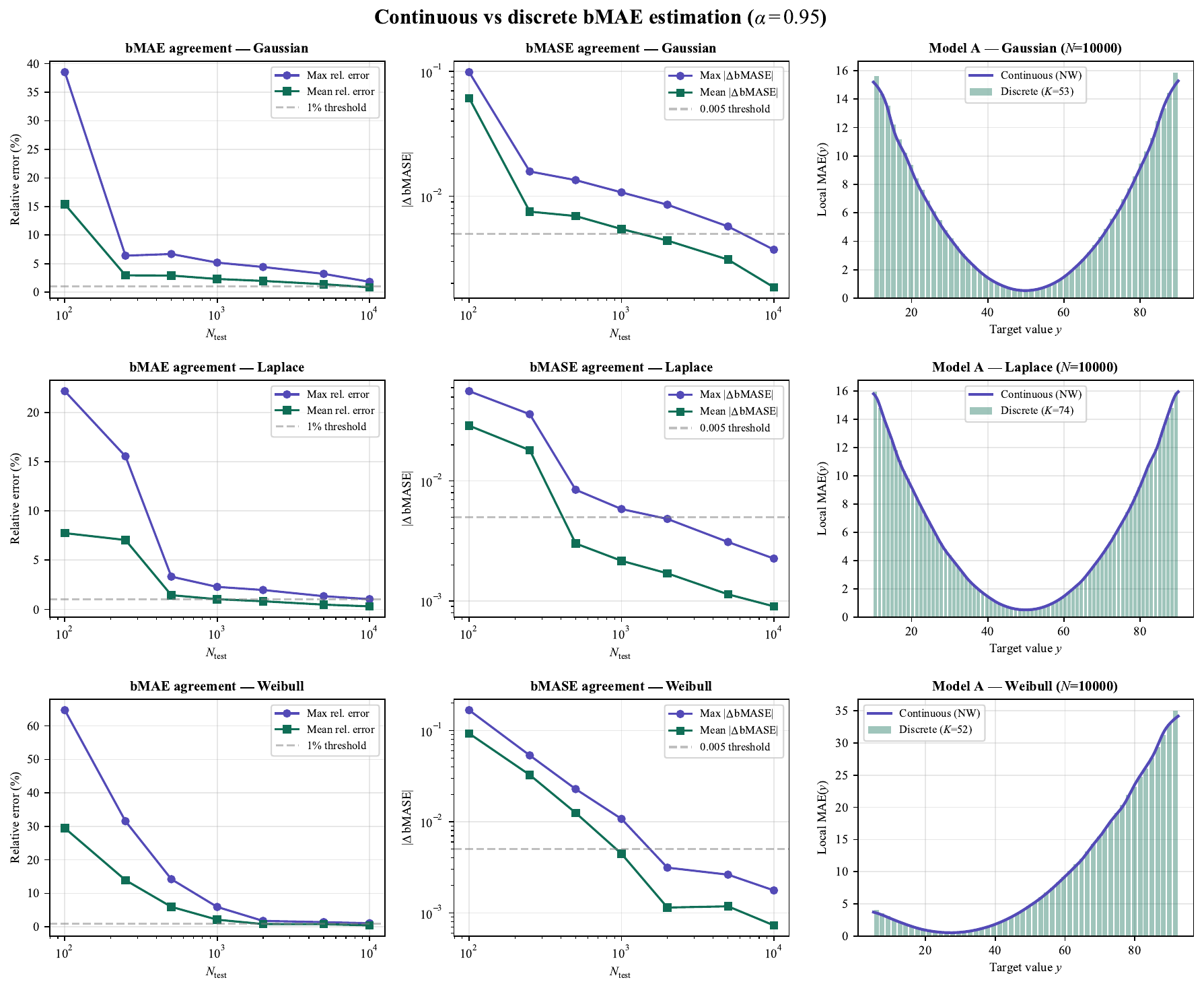}
  \caption{Agreement between the continuous (Nadaraya--Watson) and
    discrete (binned, Freedman--Diaconis) bMAE estimators across
    test-set sizes and distribution families ($\alpha=0.95$).
    \textbf{Left:} maximum relative error of the discrete estimator on
    bMAE. \textbf{Centre:} maximum absolute
    $|\Delta\,\mathrm{bMASE}|$. \textbf{Right:} local MAE curves for
    Model~A at $N_{\mathrm{test}}=10{,}000$, with the smooth NW
    estimate (blue line) overlaid on discrete bin averages (green
    bars).}
  \label{fig:cont-vs-disc}
\end{figure}

\autoref{fig:cont-vs-disc} compares the two estimators across seven test-set sizes and three distribution families at $\alpha=0.95$. For $N_{\mathrm{test}}\geq 2000$, the discrete and continuous estimators agree to within $1$--$5\%$ on bMAE and to within $0.01$ on bMASE across all three families. The agreement tightens monotonically with sample size, reaching $|\Delta\,\mathrm{bMASE}|<0.004$ at $N_{\mathrm{test}}=10{,}000$. The bMASE agreement is somewhat tighter than the raw bMAE agreement because numerator and denominator share the same binning, which partially cancels $K$-induced bias in the ratio. The right column of \autoref{fig:cont-vs-disc} confirms this visually: at the largest sample size, the discrete bin averages track the smooth NW estimate of the local MAE function across the entire target range for all three distributions. We therefore adopt the discrete estimator as the default throughout this work. It is simpler to implement, transparent in its assumptions, and produces identical model rankings to the continuous estimator on all settings considered.

\paragraph{Why $L_1$ rather than squared-error balanced metrics.}
Balanced metrics macro-average errors across target bins, so bins with few samples contribute as much as densely populated bins. This makes the stability of each per-bin estimate important. For a sample-mean estimator $\frac{1}{n}\sum_i Z_i$, the variance is $\mathrm{Var}(Z_i)/n$. For MAE, $Z_i=|e_i|$, whose variance depends on $\mathbb{E}[e_i^2]$. For MSE, $Z_i=e_i^2$, whose variance depends on $\mathbb{E}[e_i^4]$. Squared-error estimates are therefore more sensitive to heavy-tailed errors and outliers, especially when $n_k$ is small~\citep{huber_robust_1964}. This motivates our use of bMASE as an $L_1$-based balanced metric.

\section{Stability Blind Spot Appendix}
\label{sec:noise_analysis}

In addition to the stability analysis based on target imbalance, we characterise the label noise present in each \textsc{MuViS} task, because label noise and target imbalance interact and confound evaluation. In low-density regions the empirical conditional expectation $\mathbb{E}[y \mid x]$ is estimated with higher variance. In sensor-based domains this interaction is compounded by physics: extreme operating conditions often coincide with higher measurement uncertainty due to sensor saturation or nonlinear transduction (e.g.\ IEC~60584-1 thermocouple tolerances scale with the measured value). Characterising the noise landscape is therefore a prerequisite for interpreting DIR results: methods that up-weight rare samples simultaneously amplify the unreliable gradients those samples carry~\citep{puetz_deconstructing_2026}.

\begin{figure}[H]
\centering
\includegraphics[width=\textwidth]{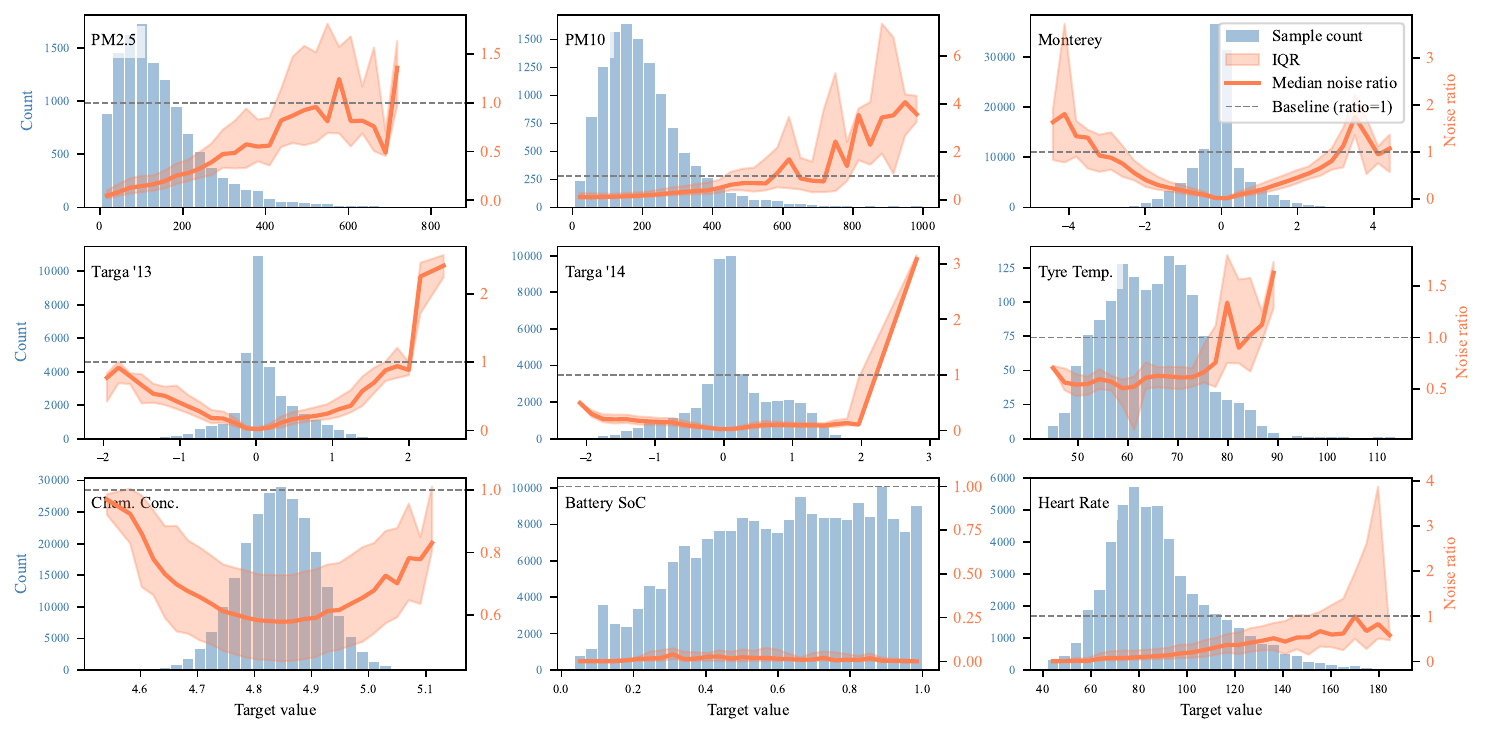}
\caption{Normalised $k$-NN noise ratio (\autoref{eq:noise_ratio}) as a
function of target value for all datasets in \textsc{MuViS}.
Blue bars: sample count per bin (left axis). Orange line and band:
median noise ratio with IQR (right axis). Dashed line: random
baseline ($r\!=\!1$).}
\label{fig:noise_profile}
\end{figure}

\paragraph{Method.}
For each training sample $(x_i, y_i)$ we identify its $k\!=\!20$ nearest neighbours in PCA-reduced feature space and compute the
\emph{normalised noise ratio}
\begin{equation}
  r_i \;=\;
  \frac{\mathrm{Var}\bigl(\{y_j : j \in \mathcal{N}_k(i)\}\bigr)}
       {\sigma^{2}(y)}\,,
  \label{eq:noise_ratio}
\end{equation}
where $\mathcal{N}_k(i)$ is the index set of the $k$ nearest neighbours of $x_i$ and $\sigma^{2}(y)$ is the global variance of the training targets. A ratio $r\!\approx\!0$ indicates a clean, locally deterministic mapping; $r\!\approx\!1$ means that local target disagreement matches the global spread; $r\!>\!1$ signals a multi-valued or strongly heteroscedastic regime. We bin per-sample ratios across the target range and report the median and IQR per bin (\autoref{fig:noise_profile}).

\paragraph{Results.}
The noise profiles reveal systematic co-occurrence of imbalance and noise (\autoref{fig:noise_profile}). The severely imbalanced tasks exhibit the most pronounced tail noise: Monterey shows a U-shaped profile rising from $r\!\approx\!0.05$ in the centre to $r\!>\!1$ in both tails, reflecting the multi-valued physics of near-limit handling; PM\textsubscript{10} displays an asymmetric pattern with clean lower values
($r\!\approx\!0.1$) but extreme upper-tail noise ($r\!>\!3.5$), consistent with sporadic, meteorologically driven pollution episodes; the Chem. Conc. shows a milder U-shape that remains below $r\!=\!1$. The moderately imbalanced tasks (Hart Rate, Tire Temp.) exhibit more uniform profiles, while battery SoC maintains $r\!<\!0.02$ throughout, confirming an almost deterministic mapping.

This pattern carries a direct methodological consequence: DIR methods that up-weight the tails of the severely imbalanced tasks must contend with amplified gradient noise alongside data scarcity, whereas on near-uniform tasks like battery SoC any tail error is purely a capacity problem, which could be the ideal scenario for DIR corrections.
The noise profiles thus refine the imbalance-only view of \autoref{tab:dataset_stats}: imbalance ratio and effective support quantify \emph{how} skewed the target distribution is, while \autoref{fig:noise_profile} quantifies how much of the residual tail-region error a model can in principle eliminate. Together, the two characterisations explain why uniform improvements across all \textsc{MuViS} tasks are difficult to achieve: the tasks for which DIR methods are most needed are also the tasks on which their corrections are most likely to amplify noise rather than correct bias.

\section{Implementation and Reproducibility}
\label{app:repro}

This section summarizes the implementation choices required to reproduce the experiments in \autoref{sec:4} alongside with the public available code at \url{www.github.com/noah-puetz/muvis-dir}. All methods use the same datasets, train/test splits, ResNet1D backbone family, evaluation metrics, and seed protocol unless stated otherwise. All experiments were implemented in Python~3.13.7 and executed on a single NVIDIA H100 (80~GB) GPU with CUDA~13.2.

\subsection{Data, Splits, and Preprocessing}
\label{app:repro:data}

All experiments use the nine \textsc{MuViS-DIR} datasets introduced in the main text. The preprocessed train/test files are inherited from \citet{brandt_muvis_2026}; the DIR experiments do not introduce additional upstream preprocessing or target transformations. Targets are predicted in their original physical units.

For each random seed, the provided training split is further divided into a training and validation subset using a stratified random 90/10 split. The test set remains fixed across all seeds and is used only for final evaluation. Input features are standardized channel-wise using statistics estimated on the training subset only; the same transformation is then applied to validation and test inputs. Targets are not standardized.

\subsection{Backbone and Training Protocol}
\label{app:repro:training}

All methods are built on the ResNet1D architecture selected in the original \textsc{MuViS} hyperparameter optimization. We keep the dataset-specific architecture parameters fixed and vary only the DIR method and its associated hyperparameters. The retained architecture parameters are shown in \autoref{tab:repro_backbone}.

Unless stated otherwise, models are trained with an $L_1$ regression loss, Adam optimization, no weight decay, no gradient clipping, and a stepwise learning-rate decay by a factor of $0.1$ at epochs 60 and 80. Training runs for 120 epochs, and the checkpoint with the lowest validation MAE is restored before test evaluation. This shared protocol applies to Vanilla, LDS, SQInv, Focal-$L_1$, and ConR. UVote and RnC use method-specific schedules, described below.

\begin{table}[t]
\centering
\caption{Dataset-specific ResNet1D backbone used in the Section~4 experiments.
The architecture parameters are inherited from the original \textsc{MuViS}
ResNet1D configuration. Here, $T$ denotes sequence length and $C$ the number
of input channels.}
\label{tab:repro_backbone}
\resizebox{\textwidth}{!}{%
\begin{tabular}{lrrrrrrl}
\toprule
Dataset & Filters & FC units & Blocks & Dropout 1 & Dropout 2 & $T$ & Target \\
\midrule
PM10                         & 156 & 512 & 4 & 0.154 & 0.028   & 24  & PM10 \\
PM2.5                        & 256 & 64  & 4 & 0.030 & 0.060   & 24  & PM2.5 \\
Heart Rate                   & 256 & 512 & 4 & 0.009 & 0.004   & 512 & heart rate \\
Battery SoC                  & 256 & 128 & 2 & 0.165 & 0.00005 & 120 & state of charge \\
Chem.\ Conc.                 & 256 & 256 & 4 & 0.093 & 0.095   & 20  & chemical concentration \\
Tire Temp.                   & 32  & 512 & 4 & 0.060 & 0.014   & 50  & front-right tire temperature \\
Monterey                     & 256 & 32  & 4 & 0.036 & 0.00053 & 20  & lateral velocity \\
Targa '13                    & 256 & 32  & 4 & 0.036 & 0.00053 & 20  & lateral velocity \\
Targa '14                    & 256 & 32  & 4 & 0.036 & 0.00053 & 20  & lateral velocity \\
\bottomrule
\end{tabular}%
}
\end{table}

\subsection{DIR Method Configurations}
\label{app:repro:methods}

The evaluated DIR methods follow the formulations described in the main text and are adapted to multimodal time-series regression. LDS, SQInv, and Focal-$L_1$ modify the regression loss through target-density- or residual-dependent weighting. ConR and RnC add contrastive objectives on the learned representation. UVote replaces the single regression head with an ensemble of expert heads trained with an uncertainty-aware objective. \autoref{tab:repro_hparams} summarizes the method-specific hyperparameters considered during configuration selection.

\begin{table}[t]
\centering
\caption{Method-specific hyperparameters used for configuration selection.
Parameters not listed here follow the shared training protocol from
\autoref{app:repro:training}.}
\label{tab:repro_hparams}
\resizebox{\textwidth}{!}{%
\begin{tabular}{lll}
\toprule
Method & Hyperparameter & Values considered \\
\midrule
LDS & number of bins & $\{50, 100\}$ \\
    & smoothing kernel & Gaussian \\
    & kernel size & $\{5, 9\}$ \\
    & smoothing bandwidth & $2.0$ \\
    & reweighting rule & inverse or square-root inverse frequency \\
\midrule
SQInv & number of bins & $\{50, 100\}$ \\
\midrule
Focal-$L_1$ & $\beta$ & $\{0.01, 0.05, 0.1, 0.2, 0.5, 1.0, 2.0, 5.0, 10.0\}$ \\
            & $\gamma$ & $\{1, 2\}$ \\
\midrule
ConR & contrastive weight & $\{0.5, 1.0, 2.0, 4.0\}$ \\
     & relative label window & $\{0.01, 0.02, 0.05, 0.10\}$ of the training target range \\
     & temperature & $0.2$ \\
\midrule
RnC & contrastive temperature & $\{0.1, 2.0\}$ \\
    & stage-1 learning rate & $\{0.05, 0.1, 0.3, 0.5\}$ \\
    & stage-1 epochs & 400 \\
    & stage-2 epochs & 90 \\
\midrule
UVote & number of experts & $\{2, 3\}$ \\
      & training schedule & 90 epochs, batch size 64 \\
\bottomrule
\end{tabular}%
}
\end{table}

\subsection{Hyperparameter Selection}
\label{app:repro:selection}

Hyperparameters are selected separately for each dataset--method pair. The selection is performed at seed 42 before the final multi-seed evaluation. For methods with a small number of predefined variants, both variants are trained and compared. For Focal-$L_1$, ConR, and RnC, we additionally run the method-specific sweeps listed in \autoref{tab:repro_hparams}.

The selected configuration is the one with the best validation balanced MAE at seed 42. After this selection step, the chosen configuration for each dataset--method pair is fixed and retrained across all ten seeds. After hyperparameter selection, one configuration is fixed for each dataset--method pair and reused for all ten seeds in the final evaluation. \autoref{tab:repro_selected_loss_methods} and \autoref{tab:repro_selected_contrastive_methods} report the effective method-specific parameters used in these final runs. Parameters that are shared across all datasets, such as the ResNet1D backbone, optimizer, training length, and validation-based checkpoint selection, are described in \autoref{app:repro:training} and are not repeated here. Vanilla uses the shared training protocol without additional method-specific parameters.

\begin{table}[t]
\centering
\caption{
Selected configurations for the loss- and reweighting-based DIR methods.
For LDS, $B$ denotes the number of target bins, $k$ the Gaussian smoothing
kernel size, and $r$ the reweighting rule. LDS uses Gaussian smoothing with
bandwidth $\sigma=2.0$ in all cases. For SQInv, $B$ denotes the number of
target bins. For Focal-$L_1$, $\beta$ and $\gamma$ denote the focal-loss
parameters.
}
\label{tab:repro_selected_loss_methods}
\scriptsize
\resizebox{\textwidth}{!}{%
\begin{tabular}{llll}
\toprule
Dataset & LDS & SQInv & Focal-$L_1$ \\
\midrule
PM10
& $B=50,\ k=9,\ r=\text{inverse}$
& $B=50$
& $\beta=0.01,\ \gamma=1$ \\

PM2.5
& $B=50,\ k=9,\ r=\text{inverse}$
& $B=50$
& $\beta=0.5,\ \gamma=1$ \\

Heart Rate
& $B=100,\ k=5,\ r=\text{sqrt-inv}$
& $B=100$
& $\beta=0.05,\ \gamma=1$ \\

Battery SoC
& $B=50,\ k=9,\ r=\text{inverse}$
& $B=100$
& $\beta=0.2,\ \gamma=2$ \\

Monterey
& $B=50,\ k=9,\ r=\text{inverse}$
& $B=100$
& $\beta=0.2,\ \gamma=2$ \\

Targa '13
& $B=50,\ k=9,\ r=\text{inverse}$
& $B=50$
& $\beta=0.2,\ \gamma=2$ \\

Targa '14
& $B=100,\ k=5,\ r=\text{sqrt-inv}$
& $B=100$
& $\beta=5.0,\ \gamma=1$ \\

Chem.\ Conc.
& $B=100,\ k=5,\ r=\text{sqrt-inv}$
& $B=100$
& $\beta=0.5,\ \gamma=1$ \\

Tire Temp.
& $B=50,\ k=9,\ r=\text{inverse}$
& $B=100$
& $\beta=0.5,\ \gamma=1$ \\
\bottomrule
\end{tabular}%
}
\end{table}

\begin{table}[t]
\centering
\caption{
Selected configurations for the contrastive and architectural DIR methods.
For ConR, $w$ denotes the target-distance window in raw target units and
$\beta$ the contrastive-loss weight; the temperature is fixed to $\tau=0.2$
and the hard-negative coefficient to $e=0.01$. For RnC, $\tau$ denotes the
contrastive temperature and $\eta_{s1}$ the stage-1 learning rate; all selected
RnC configurations use 400 contrastive pretraining epochs followed by 90
linear-probing epochs. For UVote, $K$ denotes the number of expert heads.
}
\label{tab:repro_selected_contrastive_methods}
\scriptsize
\resizebox{\textwidth}{!}{%
\begin{tabular}{llll}
\toprule
Dataset & ConR & RnC & UVote \\
\midrule
PM10
& $w=1.0,\ \beta=4.0$
& $\tau=2.0,\ \eta_{s1}=0.1$
& $K=3$ \\

PM2.5
& $w=2.0,\ \beta=2.0$
& $\tau=2.0,\ \eta_{s1}=0.5$
& $K=3$ \\

Heart Rate
& $w=1.0,\ \beta=2.0$
& $\tau=2.0,\ \eta_{s1}=0.1$
& $K=2$ \\

Battery SoC
& $w=1.0,\ \beta=4.0$
& $\tau=2.0,\ \eta_{s1}=0.3$
& $K=2$ \\

Monterey
& $w=1.0,\ \beta=0.5$
& $\tau=2.0,\ \eta_{s1}=0.5$
& $K=3$ \\

Targa '13
& $w=1.0,\ \beta=4.0$
& $\tau=2.0,\ \eta_{s1}=0.5$
& $K=3$ \\

Targa '14
& $w=1.0,\ \beta=2.0$
& $\tau=2.0,\ \eta_{s1}=0.1$
& $K=3$ \\

Chem.\ Conc.
& $w=1.0,\ \beta=4.0$
& $\tau=2.0,\ \eta_{s1}=0.3$
& $K=3$ \\

Tire Temp.
& $w=2.0,\ \beta=2.0$
& $\tau=2.0,\ \eta_{s1}=0.05$
& $K=3$ \\
\bottomrule
\end{tabular}%
}
\end{table}

\subsection{Seeds and Statistical Reporting}
\label{app:repro:seeds}

The final experiments are run with ten seeds:
\[
\{42,43,44,45,46,47,48,49,50,51\}.
\]
Each seed controls the train/validation split, model initialization, data
shuffling, dropout, and worker-level randomness in the data-loading pipeline.
We also set deterministic backend options where available. However, strict
bitwise determinism is not enforced, since some GPU operations may remain
implementation-dependent. The reported results should therefore be interpreted
as statistically reproducible rather than guaranteed bitwise identical across
all hardware and software stacks.

All tables in \autoref{sec:4} report the mean across the ten seeds. Confidence
intervals are computed from the seed-wise results using the procedure
described in the main text. The final multi-seed evaluation consists of
$$
9 \text{ datasets} \times 7 \text{ methods} \times 10 \text{ seeds}
= 630
$$
training runs.

\section{Experiments Appendix}
\label{app:experiments}

\subsection{Re-evaluation of \textsc{MuViS} Models}
\label{app:reevaluation}

To contextualize the proposed \textsc{MuViS-DIR} results, we re-evaluate the six original \textsc{MuViS} baselines under the balanced metrics introduced in \autoref{sec:3_2}. All results are averaged across ten random seeds and use the original training protocol of \citet{brandt_muvis_2026}. In contrast to the DIR experiments in \autoref{sec:4}, these models were trained with an MSE loss rather than an $L_1$ loss. Differences between the ResNet1D results in \autoref{tab:app_bmae} and \autoref{tab:app_bmase} and the corresponding results in \autoref{tab:dir-results} and \autoref{tab:bmae-vs-mae} therefore reflect the change in training loss. This comparison provides additional context on how the choice of optimization objective affects balanced and tail-sensitive performance.

\begin{table}[htbp]
\centering
\caption{Balanced Mean Absolute Error (bMAE) $\pm$ standard deviation across seeds per dataset and model.}
\label{tab:app_bmae}
\resizebox{\textwidth}{!}{
\begin{tabular}{lcccccc}
\toprule
Dataset & CatBoost & LSTM & MLP & ResNet1D & Transformer & XGBoost \\
\midrule
PM10 & 230.412 ± 2.653 & 210.981 ± 11.933 & 210.117 ± 4.624 & 226.208 ± 6.264 & 238.499 ± 5.879 & 229.785 ± 1.310 \\
PM2.5 & 172.316 ± 2.501 & 158.011 ± 8.883 & 164.419 ± 10.582 & 170.355 ± 6.330 & 177.748 ± 3.705 & 167.499 ± 3.115 \\
Heart Rate & 11.120 ± 0.080 & 9.182 ± 1.368 & 11.348 ± 0.321 & 3.279 ± 0.279 & 3.542 ± 0.164 & 12.437 ± 0.083 \\
Battery SoC & 0.008 ± 0.000 & 0.006 ± 0.000 & 0.006 ± 0.000 & 0.006 ± 0.000 & 0.007 ± 0.000 & 0.017 ± 0.000 \\
Monterey & 0.194 ± 0.005 & 0.157 ± 0.004 & 0.185 ± 0.005 & 0.158 ± 0.007 & 0.227 ± 0.010 & 0.230 ± 0.004 \\
Targa '13 & 0.150 ± 0.003 & 0.089 ± 0.003 & 0.126 ± 0.004 & 0.096 ± 0.009 & 0.198 ± 0.016 & 0.182 ± 0.003 \\
Targa '14 & 0.063 ± 0.001 & 0.065 ± 0.002 & 0.070 ± 0.002 & 0.062 ± 0.003 & 0.081 ± 0.004 & 0.061 ± 0.001 \\
Chem.\ Conc. & 0.082 ± 0.000 & 0.083 ± 0.001 & 0.082 ± 0.002 & 0.082 ± 0.001 & 0.088 ± 0.002 & 0.082 ± 0.000 \\
Tire Temp. & 3.817 ± 0.091 & 2.600 ± 0.655 & 2.617 ± 0.267 & 3.225 ± 0.499 & 3.735 ± 0.615 & 3.618 ± 0.106 \\
\bottomrule
\end{tabular}
}
\end{table}

\begin{table}[htbp]
\centering
\caption{Balanced Mean Absolute Scaled Error (bMASE) $\pm$ standard deviation across seeds per dataset and model.}
\label{tab:app_bmase}
\resizebox{\textwidth}{!}{
\begin{tabular}{lccccccc}
\toprule
Dataset & CatBoost & LSTM & MLP & ResNet1D & Transformer & XGBoost & GMean (DATA) \\
\midrule
PM10 & 0.654 ± 0.008 & 0.599 ± 0.034 & 0.596 ± 0.013 & 0.642 ± 0.018 & 0.677 ± 0.017 & 0.652 ± 0.004 & 0.636 \\
PM2.5 & 0.530 ± 0.008 & 0.486 ± 0.027 & 0.506 ± 0.033 & 0.524 ± 0.019 & 0.547 ± 0.011 & 0.516 ± 0.010 & 0.518 \\
Heart Rate & 0.275 ± 0.002 & 0.227 ± 0.034 & 0.281 ± 0.008 & 0.081 ± 0.007 & 0.088 ± 0.004 & 0.308 ± 0.002 & 0.184 \\
Battery SoC & 0.032 ± 0.001 & 0.024 ± 0.001 & 0.026 ± 0.001 & 0.024 ± 0.001 & 0.029 ± 0.001 & 0.071 ± 0.001 & 0.032 \\
Monterey & 0.111 ± 0.003 & 0.090 ± 0.003 & 0.106 ± 0.003 & 0.091 ± 0.004 & 0.130 ± 0.006 & 0.132 ± 0.002 & 0.109 \\
Targa '13 & 0.176 ± 0.004 & 0.105 ± 0.003 & 0.148 ± 0.005 & 0.113 ± 0.011 & 0.233 ± 0.019 & 0.214 ± 0.004 & 0.158\\
Targa '14 & 0.103 ± 0.001 & 0.107 ± 0.004 & 0.114 ± 0.003 & 0.102 ± 0.004 & 0.134 ± 0.006 & 0.100 ± 0.002 & 0.109 \\
Chem.\ Conc. & 0.606 ± 0.001 & 0.618 ± 0.009 & 0.611 ± 0.015 & 0.606 ± 0.006 & 0.653 ± 0.012 & 0.609 ± 0.001 & 0.617 \\
Tire Temp. & 0.407 ± 0.009 & 0.278 ± 0.070 & 0.279 ± 0.029 & 0.344 ± 0.053 & 0.399 ± 0.066 & 0.386 ± 0.011 & 0.345\\
\midrule
\textbf{GMean (MODEL)} & 0.226 & 0.187 & 0.206 & \textbf{0.174} & 0.215 & 0.257 \\
\bottomrule
\end{tabular}
}
\end{table}

\begin{table}[t]  
\centering  
\caption{  
Mean bMASE {\scriptsize{[CI low, CI high]}} across ten seeds for each method--dataset pair. Values below \textsc{Vanilla} are italicized; the best value in each row is bold. GMean aggregates across datasets via geometric mean. Lower is better; $\mathrm{bMASE}<1$ improves over the trivial median predictor.  
}  

\resizebox{\textwidth}{!}{
\begin{tabular}{lrrrrrrr}
\toprule
Dataset & Vanilla & ConR & Focal-$L_1$ & LDS & RnC & SQInv & UVote \\
\midrule
Battery SoC & 0.0229\scriptsize{ [0.0226, 0.0233]} & 0.0306\scriptsize{ [0.0231, 0.0410]} & 0.0280\scriptsize{ [0.0238, 0.0339]} & \textbf{\textit{0.0224}}\scriptsize{ [0.0221, 0.0228]} & 0.2536\scriptsize{ [0.1614, 0.3568]} & 0.0262\scriptsize{ [0.0226, 0.0331]} & 0.0251\scriptsize{ [0.0242, 0.0259]} \\
Chem. Conc. & 0.6910\scriptsize{ [0.6303, 0.7721]} & \textit{0.6485}\scriptsize{ [0.6154, 0.7120]} & \textit{0.6343}\scriptsize{ [0.6230, 0.6498]} & \textbf{\textit{0.5826}}\scriptsize{ [0.5547, 0.6197]} & \textit{0.6349}\scriptsize{ [0.6165, 0.6558]} & \textit{0.6371}\scriptsize{ [0.5861, 0.7207]} & \textit{0.6056}\scriptsize{ [0.6026, 0.6082]} \\
Heart Rate & 0.1277\scriptsize{ [0.1237, 0.1313]} & 0.1925\scriptsize{ [0.1210, 0.3051]} & \textbf{\textit{0.0748}}\scriptsize{ [0.0723, 0.0778]} & \textit{0.1130}\scriptsize{ [0.1074, 0.1222]} & \textit{0.0923}\scriptsize{ [0.0871, 0.0975]} & \textit{0.1232}\scriptsize{ [0.1067, 0.1535]} & \textit{0.0782}\scriptsize{ [0.0745, 0.0823]} \\
Monterey & 0.2511\scriptsize{ [0.1502, 0.3557]} & 0.2695\scriptsize{ [0.1536, 0.3860]} & \textit{0.2215}\scriptsize{ [0.1472, 0.2972]} & \textit{0.2401}\scriptsize{ [0.1444, 0.3368]} & 0.3622\scriptsize{ [0.1422, 0.6685]} & \textit{0.2260}\scriptsize{ [0.1390, 0.3204]} & \textbf{\textit{0.0853}}\scriptsize{ [0.0838, 0.0870]} \\
PM10 & 0.6869\scriptsize{ [0.6460, 0.7425]} & \textit{0.6793}\scriptsize{ [0.6562, 0.7079]} & \textit{0.6171}\scriptsize{ [0.6080, 0.6268]} & \textbf{\textit{0.4765}}\scriptsize{ [0.4537, 0.5046]} & \textit{0.6663}\scriptsize{ [0.6576, 0.6746]} & \textit{0.6230}\scriptsize{ [0.6047, 0.6475]} & \textit{0.6792}\scriptsize{ [0.6444, 0.7150]} \\
PM2.5 & 0.5705\scriptsize{ [0.5418, 0.5991]} & \textit{0.5507}\scriptsize{ [0.5242, 0.5820]} & \textit{0.5242}\scriptsize{ [0.5129, 0.5376]} & \textit{0.5489}\scriptsize{ [0.4967, 0.6049]} & \textit{0.5675}\scriptsize{ [0.5482, 0.5937]} & \textit{0.5457}\scriptsize{ [0.5177, 0.5751]} & \textbf{\textit{0.5142}}\scriptsize{ [0.5065, 0.5225]} \\
Targa '13 & \textbf{0.1116}\scriptsize{ [0.1064, 0.1158]} & 0.1136\scriptsize{ [0.1111, 0.1161]} & 0.1933\scriptsize{ [0.1844, 0.2027]} & 0.1212\scriptsize{ [0.1175, 0.1253]} & 0.3452\scriptsize{ [0.1867, 0.5560]} & 0.1132\scriptsize{ [0.1106, 0.1163]} & 0.1136\scriptsize{ [0.1081, 0.1182]} \\
Targa '14 & 0.1276\scriptsize{ [0.0988, 0.1686]} & 0.1319\scriptsize{ [0.0974, 0.1793]} & 0.1326\scriptsize{ [0.1239, 0.1437]} & \textit{0.1273}\scriptsize{ [0.1038, 0.1604]} & 0.1746\scriptsize{ [0.1442, 0.2107]} & \textit{0.1183}\scriptsize{ [0.1044, 0.1434]} & \textbf{\textit{0.1055}}\scriptsize{ [0.0957, 0.1222]} \\
Tire Temp. & 0.3662\scriptsize{ [0.3357, 0.3949]} & \textit{0.3618}\scriptsize{ [0.3389, 0.3849]} & 0.3777\scriptsize{ [0.3482, 0.4051]} & \textit{0.3534}\scriptsize{ [0.3283, 0.3805]} & 0.3933\scriptsize{ [0.3578, 0.4295]} & \textit{0.3505}\scriptsize{ [0.3289, 0.3731]} & \textbf{\textit{0.3313}}\scriptsize{ [0.3057, 0.3565]} \\
\midrule
GMean & 0.2163\scriptsize{ [0.1903, 0.2423]} & 0.2338\scriptsize{ [0.1907, 0.2822]} & \textit{0.2135}\scriptsize{ [0.1941, 0.2330]} & \textit{0.1997}\scriptsize{ [0.1771, 0.2229]} & 0.3314\scriptsize{ [0.2532, 0.4071]} & \textit{0.2084}\scriptsize{ [0.1833, 0.2407]} & \textbf{\textit{0.1732}}\scriptsize{ [0.1657, 0.1818]} \\
\bottomrule
\end{tabular}
}
\label{tab:app_dir_bmase_ci}
\end{table}

\subsection{Performance comparison}
\label{app:performance_comparison}

The following figures complement the aggregate results in the main text by showing per-bin MAE curves for all methods across the nine \textsc{MuViS-DIR} datasets. These plots make the regional structure of the errors visible and show where individual methods improve over, match, or degrade relative to the vanilla baseline. They also illustrate that improvements in aggregate bMAE or bMASE can arise from different error profiles across the target range, motivating the use of both scalar balanced metrics and per-bin diagnostic plots.

\begin{figure}[p]
    \centering

    \begin{minipage}[t]{0.45\linewidth}
        \centering
        \includegraphics[width=\linewidth]{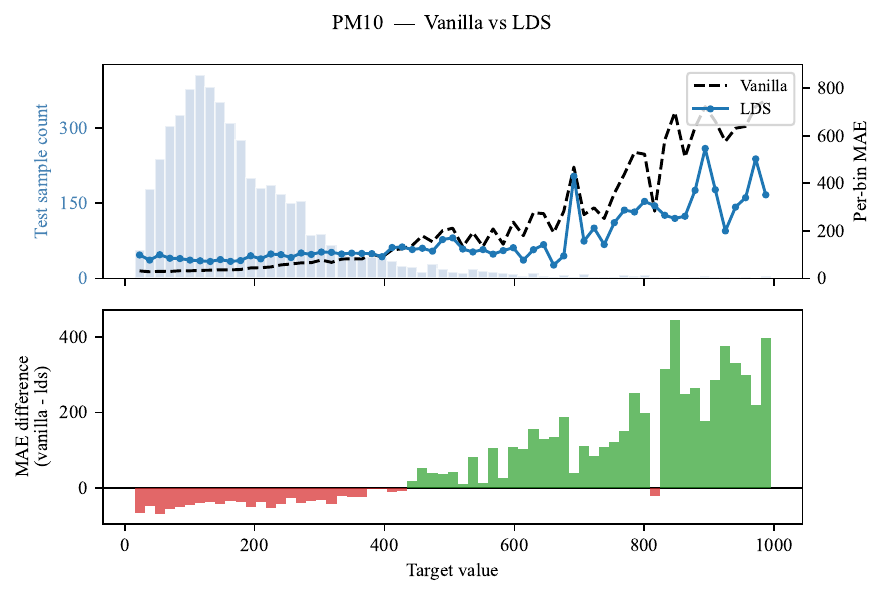}

        \vspace{0.4em}
        \includegraphics[width=\linewidth]{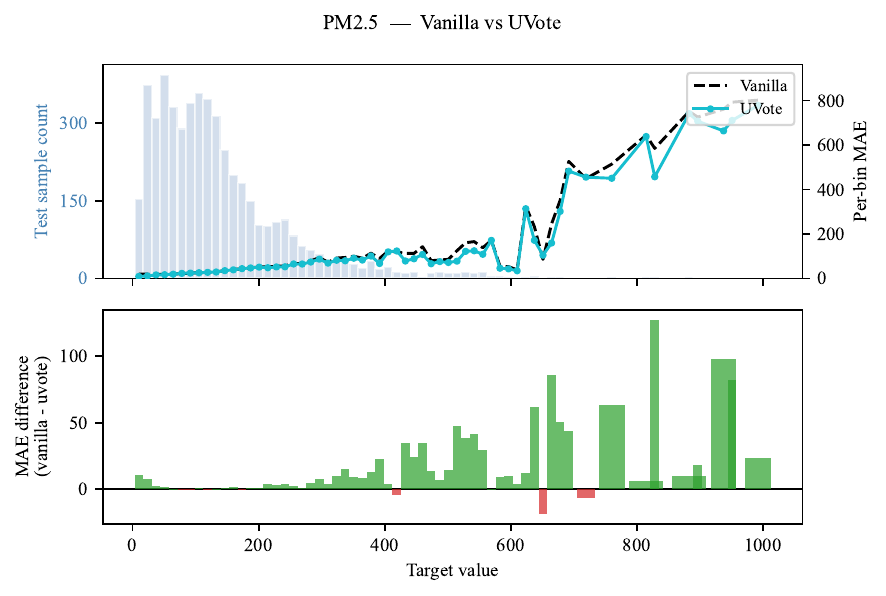}

        \vspace{0.4em}
        \includegraphics[width=\linewidth]{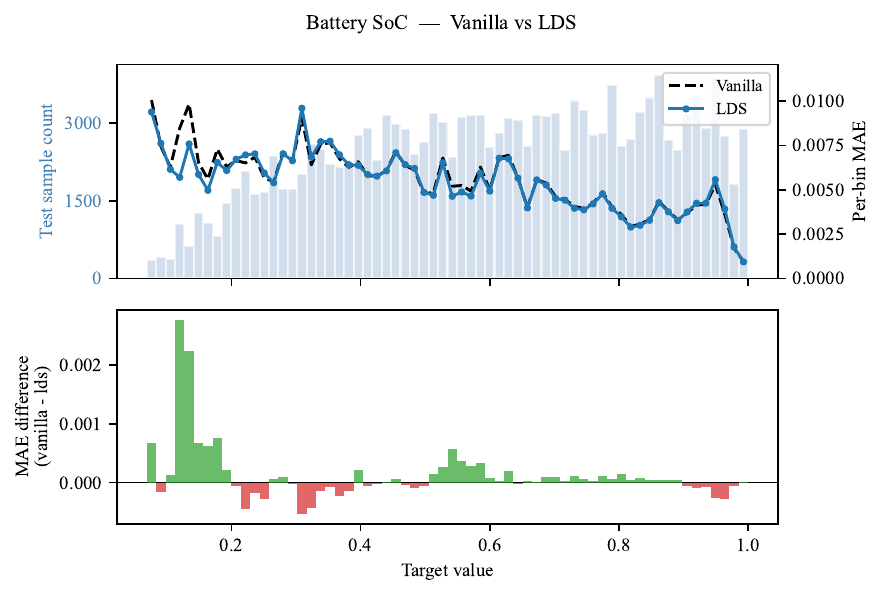}

        \vspace{0.4em}
        \includegraphics[width=\linewidth]{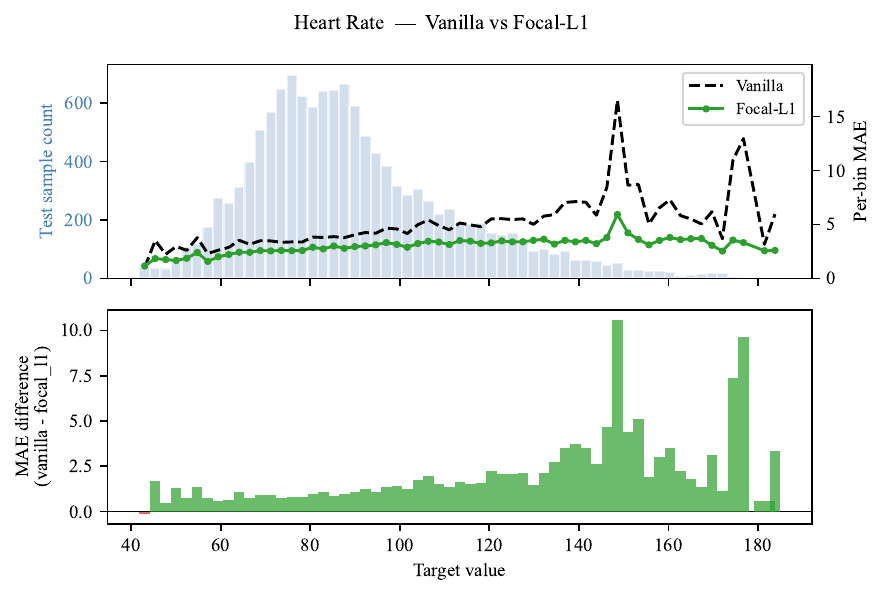}

        \vspace{0.4em}
        \includegraphics[width=\linewidth]{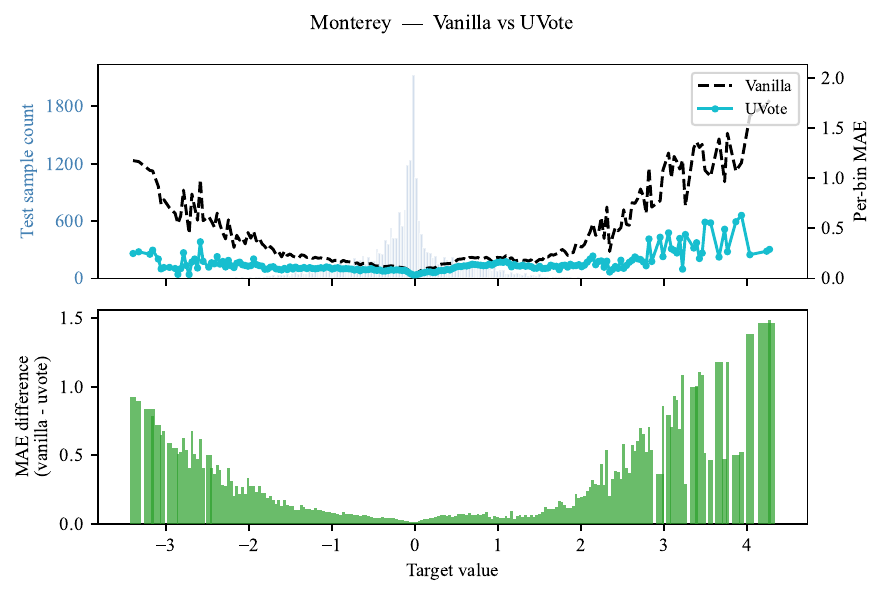}
    \end{minipage}
    \hfill
    \begin{minipage}[t]{0.45\linewidth}
        \centering
        \includegraphics[width=\linewidth]{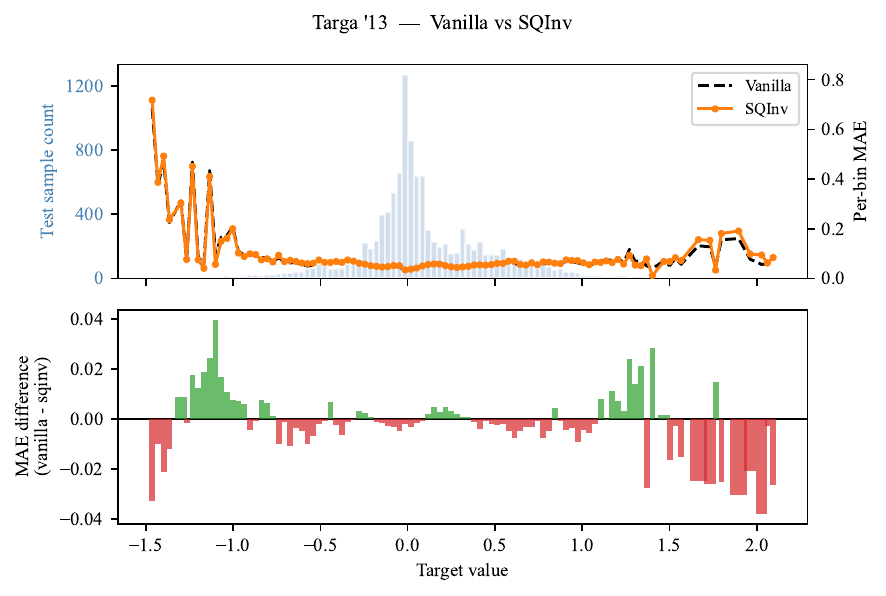}

        \vspace{0.4em}
        \includegraphics[width=\linewidth]{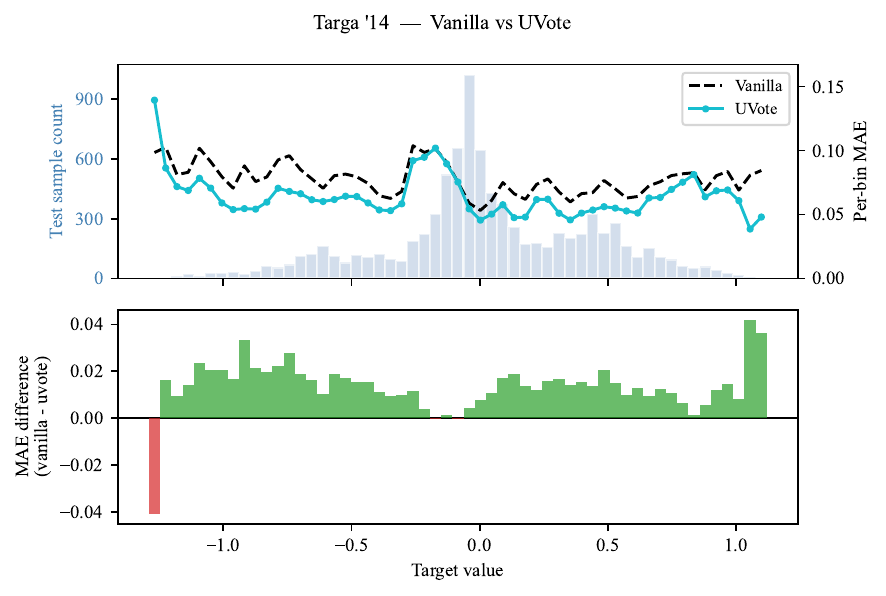}

        \vspace{0.4em}
        \includegraphics[width=\linewidth]{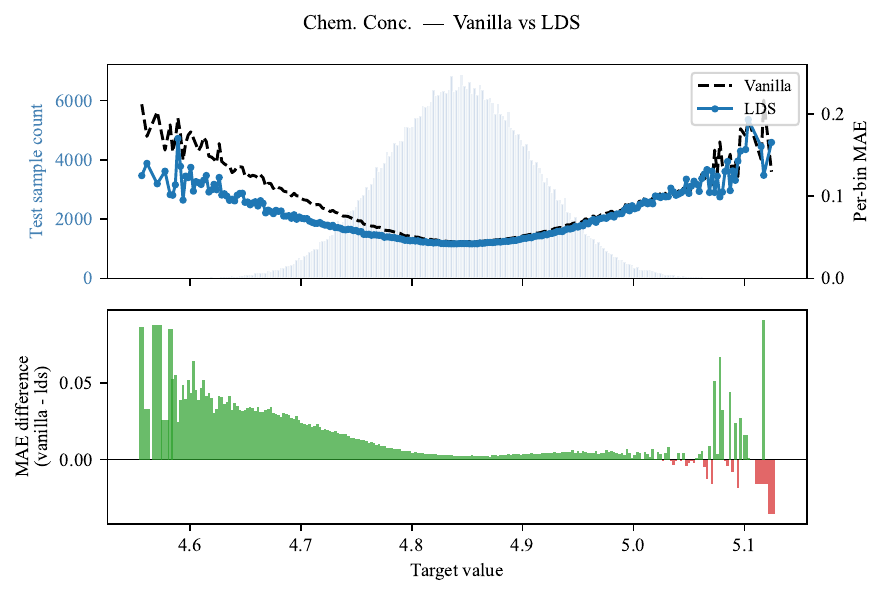}

        \vspace{0.4em}
        \includegraphics[width=\linewidth]{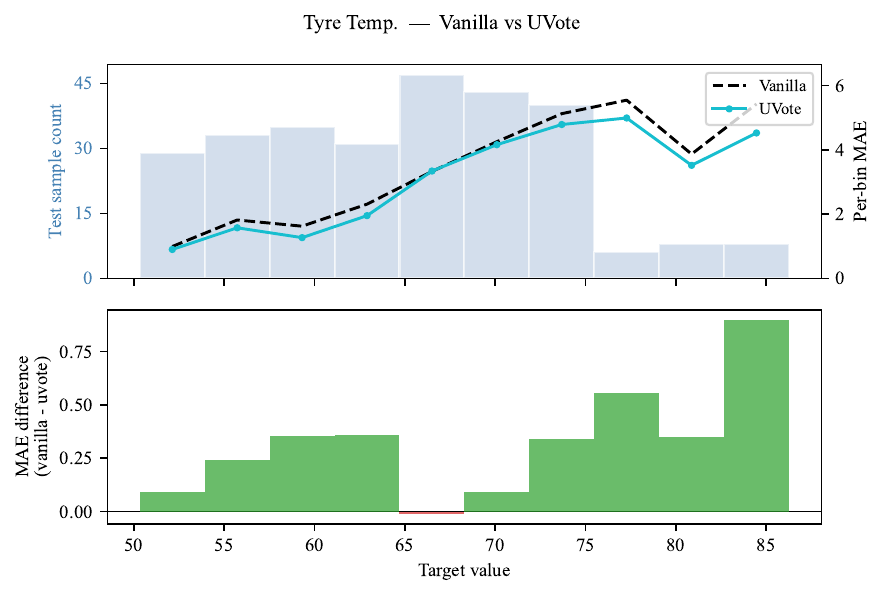}
    \end{minipage}

    \caption{Comparison of the respective best performing DIR methods against the vanilla baseline across all nine \textsc{MuViS-DIR} datasets. For every figure top row compares per-bin MAE; bottom row shows the signed difference to the baseline.}
    \label{fig:appendix_method_vs_vanilla}
\end{figure}

Finally, \autoref{fig:appendix_method_vs_vanilla} compares the best-performing DIR method for each dataset directly against the vanilla baseline. The signed difference plots highlight whether improvements are concentrated in sparse target regions or distributed more uniformly across the target range.

\begin{figure}
    \centering
    \includegraphics[width=0.9\linewidth]{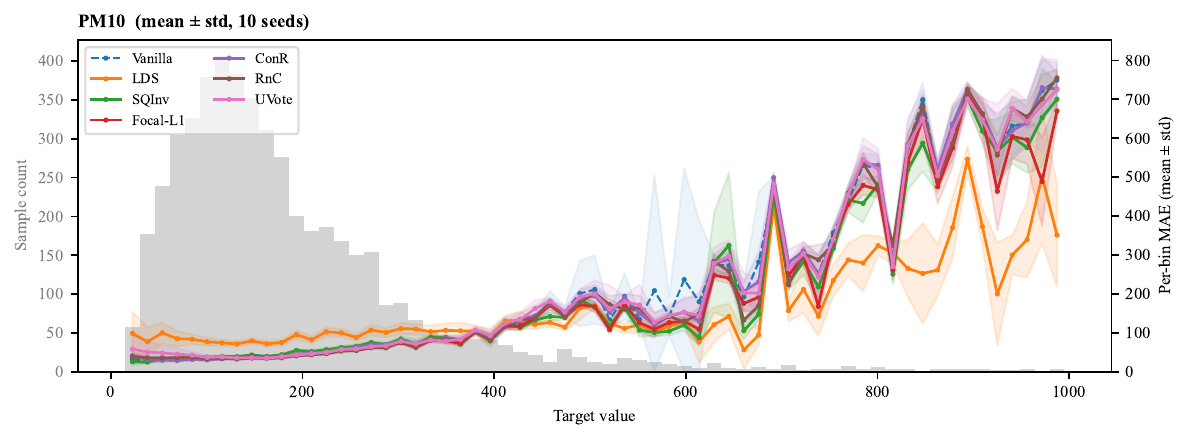}
    \includegraphics[width=0.9\linewidth]{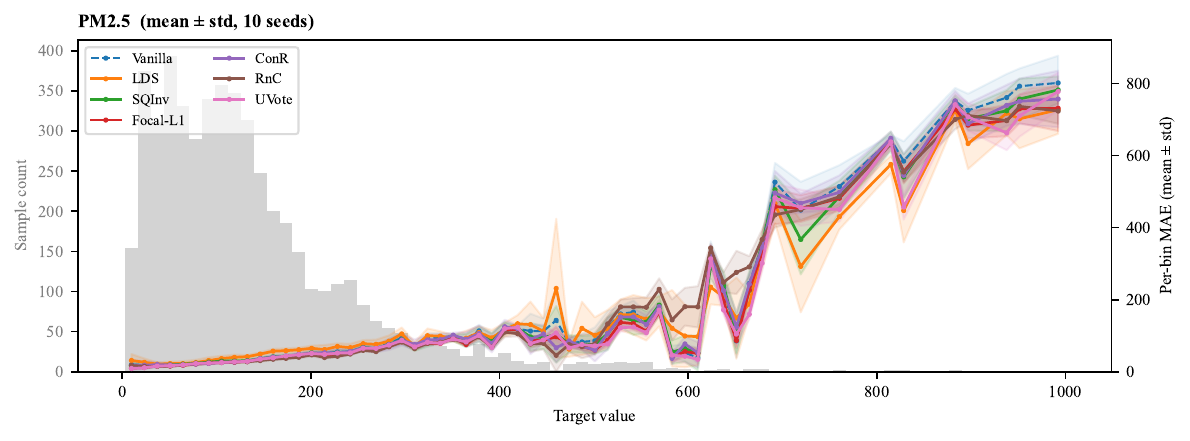}
    \includegraphics[width=0.9\linewidth]{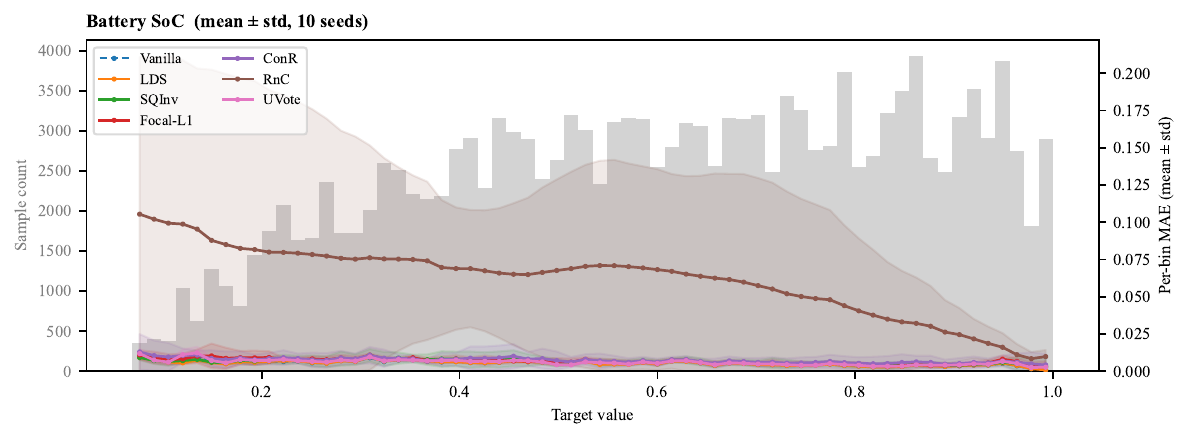}
    \caption{
    Per-bin MAE of the six representative DIR methods and the vanilla baseline across three \textsc{MuViS} tasks \citep{brandt_muvis_2026}. From top to bottom, rows show PM10, Beijing PM25, and Battery SoC prediction. The plots illustrate how method performance varies across the target range.
    }
    \label{fig:appendix_all_methods_air_battery}
\end{figure}

\begin{figure}
    \centering
    \includegraphics[width=0.9\linewidth]{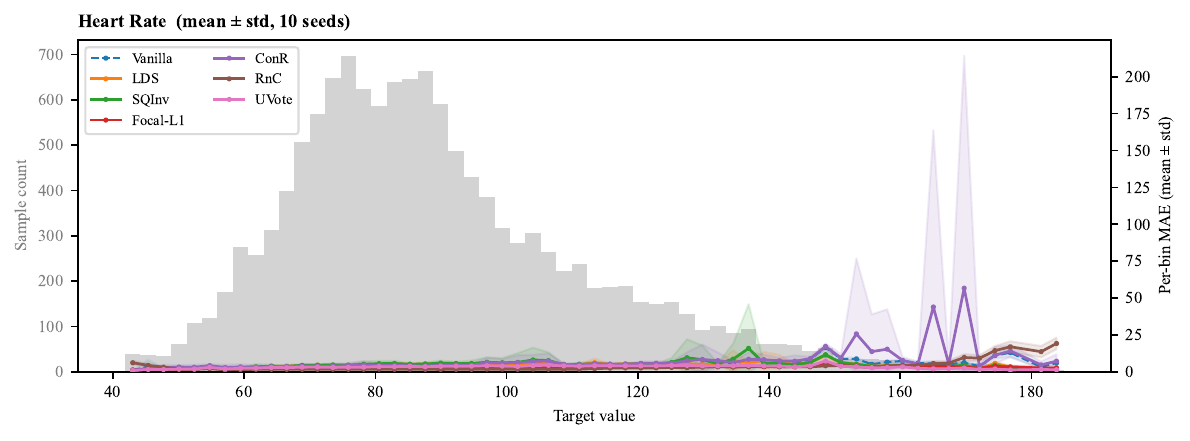}
    \includegraphics[width=0.9\linewidth]{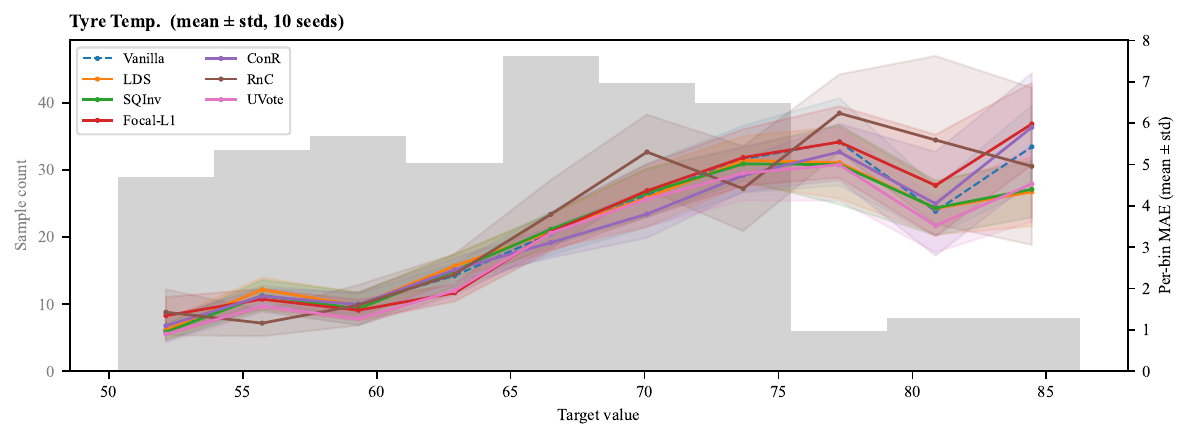}
    \includegraphics[width=0.9\linewidth]{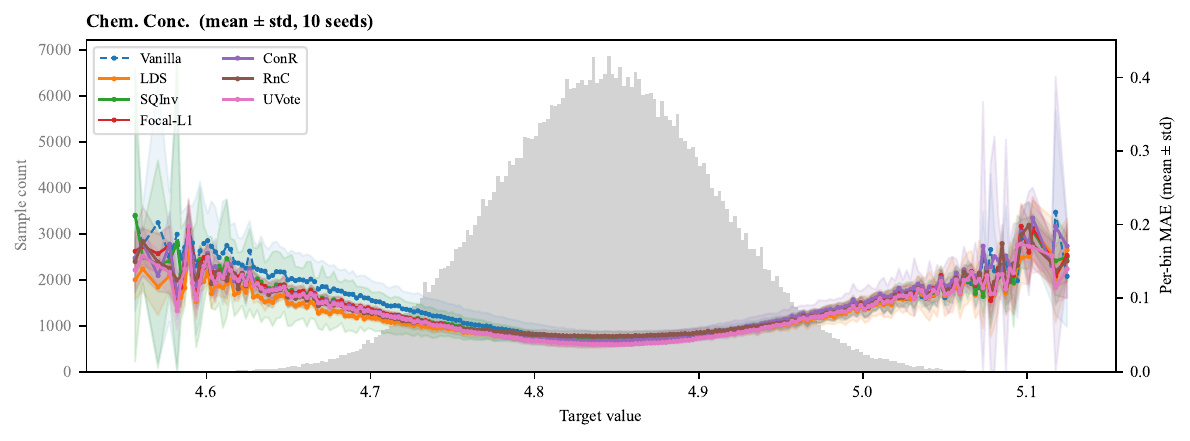}
    \caption{Per-bin MAE of the six representative DIR methods and the vanilla baseline across three \textsc{MuViS} tasks \citep{brandt_muvis_2026}. From top to bottom, rows show PPG, Tire. Temp, and Chem. Conc. prediction. The plots illustrate how method performance varies across the target range.}
    \label{fig:appendix_all_methods_bio_process_vehicle}
\end{figure}

\begin{figure}

    \includegraphics[width=0.9\linewidth]{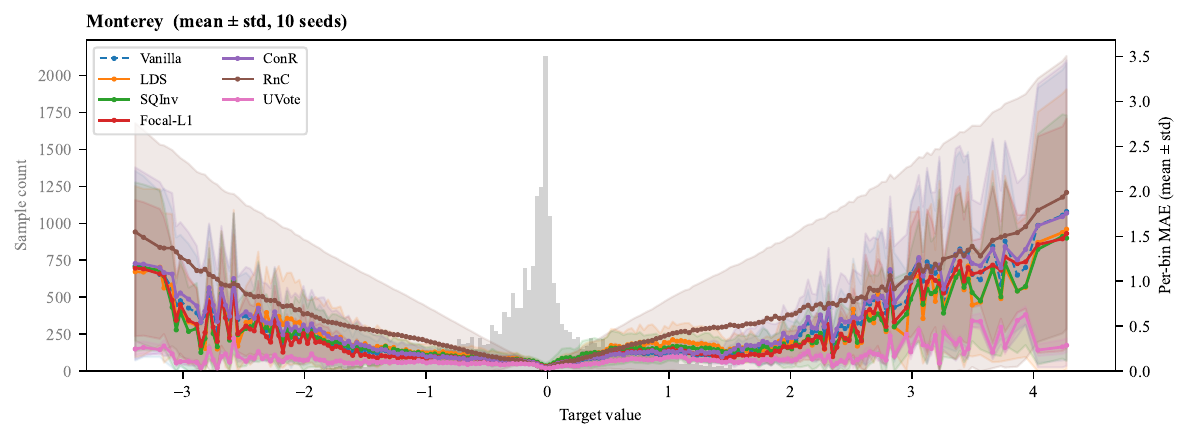}
    \includegraphics[width=0.9\linewidth]{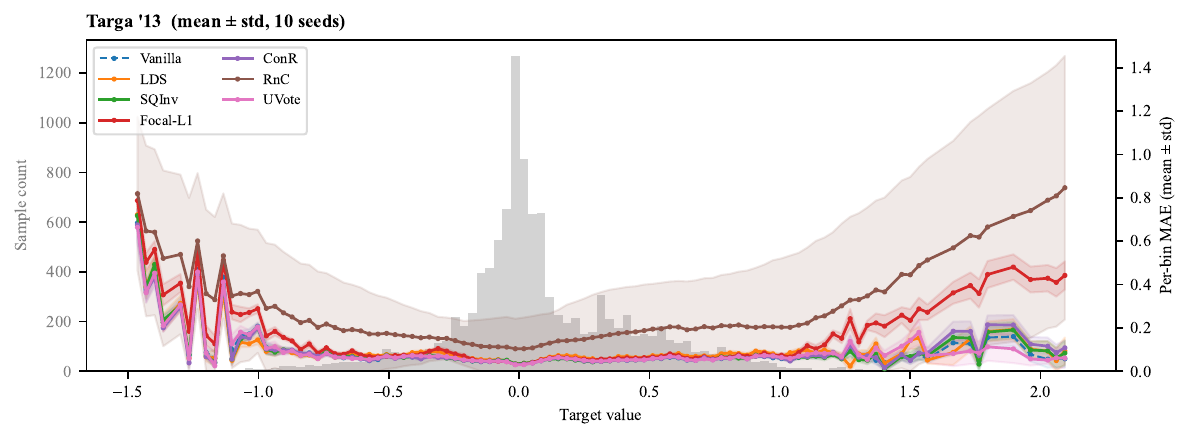}
    \includegraphics[width=0.9\linewidth]{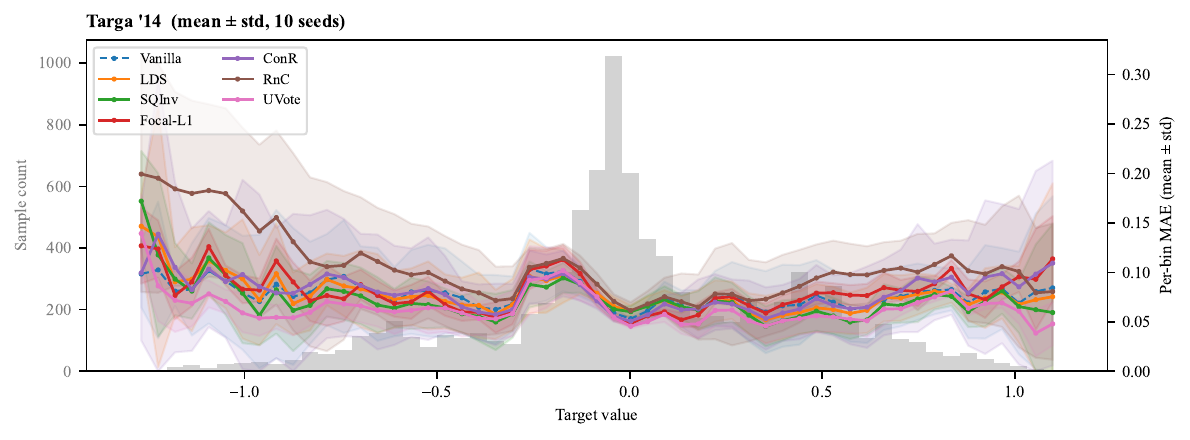}
    \caption{Per-bin MAE of the six representative DIR methods and the vanilla baseline across three \textsc{MuViS} tasks \citep{brandt_muvis_2026}. From top to bottom, rows show Monterey, Targa '13, and Targa '14 prediction. The plots illustrate how method performance varies across the target range.}
    \label{fig:appendix_all_methods_revs}
\end{figure}

\end{document}